\documentclass[journal,twoside,web]{ieeecolor}
\usepackage{etoolbox}
\makeatletter
\@ifundefined{color@begingroup}%
{\let\color@begingroup\relax
\let\color@endgroup\relax}{}%
\def\fix@ieeecolor@hbox#1{%
\hbox{\color@begingroup#1\color@endgroup}}
\patchcmd\@makecaption{\hbox}{\fix@ieeecolor@hbox}{}{\FAILED}
\patchcmd\@makecaption{\hbox}{\fix@ieeecolor@hbox}{}{\FAILED}
\definecolor{lightgray}{RGB}{237.15,237.15,237.15} 
\usepackage{tmi}
\usepackage{titlesec}
\titlespacing{\section}{1pt}{*0.5}{*0.5}
\titlespacing{\subsection}{0.5pt}{*0.5}{*0.5}
\titlespacing{\subsubsection}{0.3pt}{*0.5}{*0.5}

\usepackage{cite}
\usepackage{amsmath,amssymb,amsfonts}
\usepackage{adjustbox}
\usepackage{graphicx}
\usepackage{rotating}
\usepackage{xcolor}
\usepackage{tabularray}
\usepackage{makecell}
\usepackage[normalem]{ulem}
\usepackage{multirow}
\usepackage{colortbl}
\usepackage{booktabs}
\usepackage{textcomp}
\usepackage{hyperref}
\hypersetup{
    colorlinks=true,
    linkcolor=red,
    citecolor=blue,
    urlcolor=red
}
\def\BibTeX{{\rm B\kern-.05em{\sc i\kern-.025em b}\kern-.08em
    T\kern-.1667em\lower.7ex\hbox{E}\kern-.125emX}}
\begin{document}
\title{MSCPT: Few-shot Whole Slide Image\\ Classification with Multi-scale and\\ Context-focused Prompt Tuning}
\author{Minghao~Han,
        Linhao~Qu,
        Dingkang~Yang,
        Xukun~Zhang,
        Xiaoying~Wang,
        Lihua~Zhang,~\IEEEmembership{Member, IEEE}
\thanks{This project was funded by the National Natural Science Foundation of China 82090052. (Minghao~Han and Linhao~Qu are the co-first authors. Corresponding authors: Dingkang~Yang; Lihua~Zhang.)}
\thanks{Minghao Han, Dingkang Yang, Xukun Zhang, and Lihua Zhang are with the Academy for Engineering and Technology, Fudan University, Shanghai 200433, China. (E-mail: \{mhhan22, zhangxk21\}@m.fudan.edu.cn, \{dkyang20, lihuazhang \}@fudan.edu.cn). Linhao~Qu is with  Fudan University, Shanghai 200032, China. (E-mail: lhqu20@fudan.edu.cn). Xiaoying~Wang is with the Zhongshan Hospital, Fudan University, Shanghai 200032, China. (E-mail: xiaoyingwang@fudan.edu.cn).}}

\maketitle

\begin{abstract}
Multiple instance learning (MIL) has become a standard paradigm for the weakly supervised classification of whole slide images (WSIs). However, this paradigm relies on using a large number of labeled WSIs for training. The lack of training data and the presence of rare diseases pose significant challenges for these methods. Prompt tuning combined with pre-trained Vision-Language models (VLMs) is an effective solution to the Few-shot Weakly Supervised WSI Classification (FSWC) task. Nevertheless, applying prompt tuning methods designed for natural images to WSIs presents three significant challenges: 1) These methods fail to fully leverage the prior knowledge from the VLM's text modality; 2) They overlook the essential multi-scale and contextual information in WSIs, leading to suboptimal results; and 3) They lack exploration of instance aggregation methods. To address these problems, we propose a Multi-Scale and Context-focused Prompt Tuning (MSCPT) method for FSWC task. Specifically, MSCPT employs the frozen large language model to generate pathological visual language prior knowledge at multiple scales, guiding hierarchical prompt tuning. Additionally, we design a graph prompt tuning module to learn essential contextual information within WSI, and finally, a non-parametric cross-guided instance aggregation module has been introduced to derive the WSI-level features. Extensive experiments, visualizations, and interpretability analyses were conducted on five datasets and three downstream tasks using three VLMs, demonstrating the strong performance of our MSCPT. All codes have been made publicly accessible at \url{https://github.com/Hanminghao/MSCPT}.

\end{abstract}

\begin{IEEEkeywords}
whole slide image classification, prompt tuning, few-shot learning, multimodal.
\end{IEEEkeywords}

\section{Introduction}
\label{intro}
\begin{figure}[t!]
   \begin{center}
   % \fbox{\rule{0pt}{2in} \rule{0.9\linewidth}{0pt}}ƒ
   \includegraphics[width=1\linewidth]{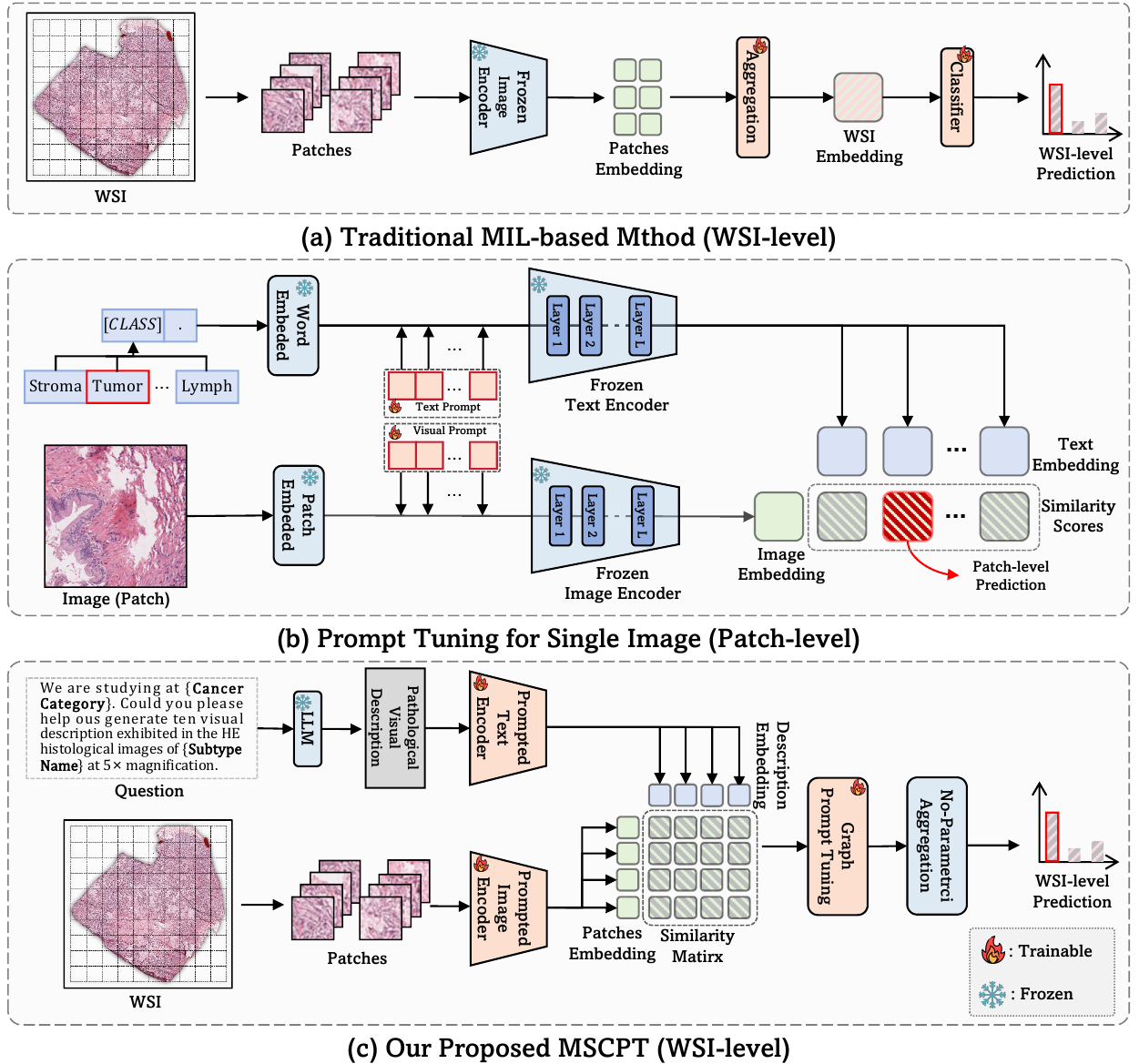}
   \end{center}
   \vspace{-0.5em}
   \caption{Motivation of our MSCPT. \textbf{(a)} Traditional MIL-based methods mainly focus on instance aggregation and require a large amount of training data. \textbf{(b)} Prompt tuning methods for natural images incorporate a set of trainable parameters into the input space for training, enabling pre-trained VLMs to be applied to downstream tasks. However, those methods are only suitable for single images and are no longer adequate for WSI-level tasks due to the enormous size of WSIs. \textbf{(c)} MSCPT leverages pathological visual descriptions combined with multimodal hierarchical prompt tuning to explore the potential of VLMs. For simplicity, we only depicted the data flow diagram for a single scale.} 
   \label{figure1}
   \vspace{-1em}
\end{figure}
Developing automated analysis frameworks using Whole Slide Images (WSIs) is crucial in clinical practice~\cite{shao2021transmil, xing2024comprehensive, guo2023predicting}, as WSIs are widely regarded as the ``gold standard" for cancer diagnosis, typing, staging, and prognosis analysis~\cite{ludwig2005biomarkers}.
Given the enormous size of WSIs (roughly 40,000 × 40,000 pixels), multiple instance learning (MIL)~\cite{ilse2020deep} has become the dominant method. As shown in Fig. \ref{figure1}\textcolor{red}{a}, traditional MIL-based methods typically follow a four-step paradigm: patch cutting, feature extraction, feature aggregation, and classification. Most MIL-based methods are conducted under weak supervision at the bag level, as creating instance-level labels is quite labor-intensive~\cite{qu2024rise, shao2023hvtsurv}. This weak supervision paradigm has led to a problem: a large number of WSIs are required to train an effective model~\cite{campanella2019clinical, qu2024rethinking}. In clinical practice, patient privacy concerns, rare diseases, and difficulties in preparing pathology slides make accumulating a large number of WSIs very challenging~\cite{srinidhi2021deep, shmatko2022artificial}.

Vision-Language models (VLMs) have shown excellent generalization ability to downstream tasks~\cite{clip, lee2020biobert, maple, blip, GexMolGen, han2024umpire}. Recently, researchers have proposed specialized VLMs for analyzing pathological images, including MI-Zero~\cite{mizero}, PLIP~\cite{plip}, and Conch~\cite{conch}. These VLMs, extensively pre-trained on abundant image-text pairs, contain significant prior knowledge. If the prior knowledge of VLMs can be fully exploited with a few training samples, it can partially alleviate the data scarcity problem in WSI classification tasks. Therefore, we aim to explore a novel ``data-efficient" method based on the VLMs to improve the model's performance on the \textbf{Few-shot Weakly Supervised WSI Classification (FSWC)}~\cite{qu2024rise} task.

Nevertheless, there is a gap between generally pre-trained VLMs and specific downstream tasks. Under the few-shot scenarios, researchers often employ prompt tuning to bridge this gap with the help of a few training samples~\cite{clip, cocoop, coop, maple}. As shown in Fig. \ref{figure1}\textcolor{red}{b}, prompt tuning aims to learn a set of trainable continuous vectors and incorporate these vectors into the input space for training, effectively adapting the fixed pre-trained VLMs for specific downstream tasks. 

However, existing prompt tuning methods for natural images (such as CoOp~\cite{coop}, CoCoOp~\cite{cocoop}, and MetaPrompt~\cite{metaprompt}) are only effective for single images (\textit{i.e.}, patch-level). Since each WSI typically contains tens of thousands of patches, these methods are ineffective for WSI-level tasks. Also, studies indicate that the multi-scale information~\cite{chen2022scaling} and the contextual information~\cite{shao2023characterizing} in WSIs play a significant role in cancer analysis, but those methods fail to capture this crucial information. Additionally, in training VLMs, the image-text pairs contain more than just information about the category. They also include more details about the image, such as contextual properties of the object~\cite{clip} and descriptions of the cellular microenvironment~\cite{plip, conch}. However, existing prompt tuning methods have primarily focused on image category information without emphasizing a detailed image content analysis, which has left the full potential of pathological VLMs underexplored.

To address the aforementioned issue, we propose \textbf{M}ulti-\textbf{S}cale and \textbf{C}ontext-focused \textbf{P}rompt \textbf{T}uning (MSCPT) for WSI classification in weakly supervised and few-shot scenarios. Our framework fully leverages the characteristic of VLM training with image-text pairs at dual magnification scales: 1) At low magnification, we provide the VLM with pathological visual descriptions at the tissue level (such as the infiltration between tumor tissue and other normal tissues); 2) At high magnification, pathological visual descriptions at the cellular level (such as cell morphology, nuclear changes, and the formation of various organelles) are provided to the VLM. These pathological visual descriptions at multiple scales can help VLM identify regions that are helpful for cancer analysis and achieve optimal results even with limited training samples.

As illustrated in Fig. \ref{figure1}\textcolor{red}{c}, the core idea behind developing MSCPT is to incorporate prior knowledge at the tissue and cellular scales into the WSI-level tasks. Specifically, we first use a frozen large language model (LLM) to generate multi-scale pathological visual descriptions as prior knowledge.

Secondly, we design a Multi-scale Hierarchical Prompt Tuning (MHPT) module to combine pathological visual descriptions from multiple scales hierarchically to enhance prompt effectiveness. Inspired by Metaprompt~\cite{metaprompt}, a dual-path asymmetric framework is adopted, asymmetrically freezing the image encoder and text encoder at different scales for prompt tuning. This asymmetric framework enables us to freeze half of the encoder to reduce the number of trainable parameters. Specifically, MHPT contains low-level and high-level prompts for both low and high-magnification visual descriptions, as well as global trainable prompts. The MHPT module employs the transformer layers in the text encoder to effectively learn the interactions among three distinct prompts.

Furthermore, the Image-text Similarity-based Graph Prompt Tuning (ISGPT) module is introduced to extract contextual information. Contextual analysis is crucial for WSI analysis~\cite{patchgcn, shao2021transmil}. Researchers typically employ Graph Neural Networks (GNNs) or Transformer architectures to learn the context of WSIs. However, due to the large size of WSIs and the substantial computational resources required by the Transformer paradigm, these methods are not suitable for the FSWC task. Therefore, MSCPT adopts GNNs for lightweight contextual learning. Specifically, we do not follow previous approaches~\cite{patchgcn, h2gt} of using patch positions or patch feature similarity to construct GNNs. Instead, we propose to use the similarity between patches and pathological visual descriptions as the basis for building GNNs. We demonstrate that image-text pairs in GNNs more effectively capture global features than methods relying on patch positions and image similarity, as confirmed by corresponding ablation experiments.

Finally, motivated by the powerful zero-shot capabilities of VLMs~\cite{plip, conch, mizero}, we propose aggregating instances based on the similarity between image patches and pathological visual descriptions. The Non-Parametric Cross-Guided Pooling (NPCGP) module, utilizing the Top-K algorithm for instance aggregation, is introduced to further reduce the risk of overfitting in few-shot scenarios and endows MSCPT with human-readable interpretability. Overall, our contributions are:
\begin{enumerate}
  \item MSCPT demonstrates that more detailed high-level concepts from pathological descriptions combined with low-level image representations can enhance few-shot weakly supervised WSI classification.
  \item MSCPT achieves excellent performance by introducing only a limited number of trainable parameters ($0.4\%$ to $0.9\%$ of the pre-trained VLM). Additionally, MSCPT is applicable to fine-tune any VLMs for WSI-level tasks.
  \item Extensive experiments, visualizations, and interpretability analyses conducted on five datasets spanning three downstream tasks and evaluated using three different VLMs confirm that MSCPT achieves state-of-the-art performance in few-shot scenarios, outperforming other traditional methods and prompt-tuning methods.
\end{enumerate}

\begin{figure*}[t!]
   \begin{center}
   % \fbox{\rule{0pt}{2in} \rule{0.9\linewidth}{0pt}}ƒ
   \includegraphics[width=1\linewidth]{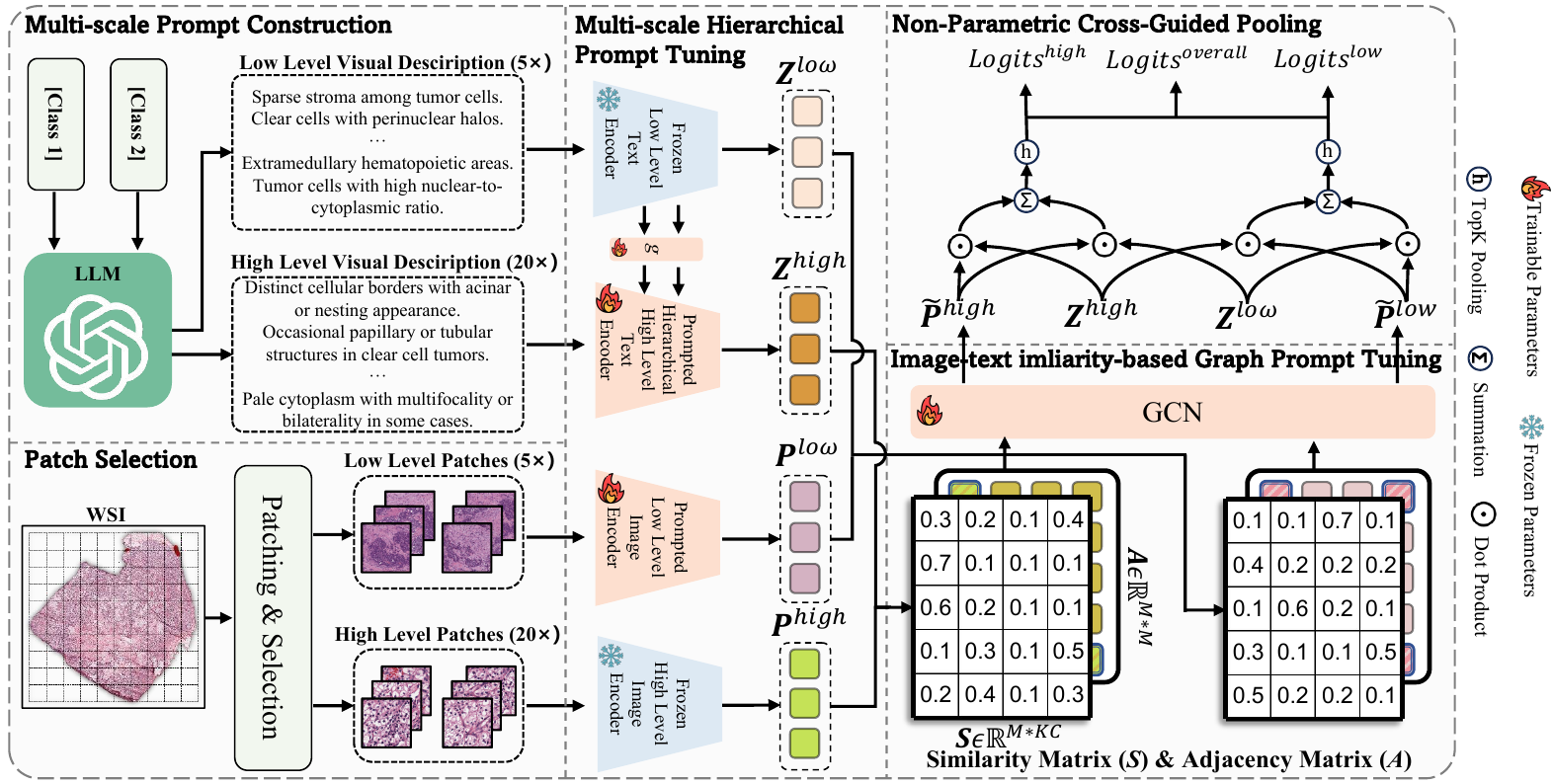}
   \end{center}
   \vspace{-0.5em}
   \caption{We develop MSCPT based on the dual-path asymmetric framework, which inputs patches and pathological visual descriptions from multi-scale to different encoders. MSCPT utilizes a large language model to generate multi-scale pathological visual descriptions. These descriptions are combined using Multi-scale Hierarchical Prompt Tuning (MHPT) to integrate information across multiple scales. Then, Image-text similarity-based Graph Prompt Tuning (ISGPT) is employed to learn context information at each scale. Finally, Non-Parametric Cross-Guided Pooling (NPCGP) aggregates instances guided by pathological visual descriptions to achieve the final Whole Slide Image classification result.}
   \label{figure2}
   \vspace{-1em}
\end{figure*}

\section{Related Work}
\subsection{Multiple Instance Learning in Whole Slide Images}
Due to the high resolution of Whole Slide Images (WSIs) and the challenges of detailed labeling, weakly supervised methods based on Multiple Instance Learning (MIL) have emerged as the mainstream for WSI analysis. The MIL-based methods treat a WSI as a bag and all patches as instances, considering a bag positive if it contains at least one positive instance. Within the MIL framework, a step is required to aggregate all instances into bag features. The most primitive aggregation methods are non-parametric mean pooling and max pooling. However, since disease-related instances are a small fraction~\cite{clam}, those non-parametric aggregation methods treated all instances equally, causing useful information to be overwhelmed by irrelevant data. Subsequently, some attention-based methods (such as ABMIL~\cite{abmil}, DSMIL~\cite{dsmil}, and CLAM~\cite{clam}) were introduced, assigning different weights to each instance and aggregating them based on the weights. Furthermore, MIL methods based on Graph Neural Networks (GNNs)~\cite{patchgcn, h2gt} and Transformers~\cite{shao2021transmil, gtmil} have also been proposed to capture both local and global contextual information of WSIs. Those methods have shown significant improvements in recent years. Still, the cost of enhancing model performance is the increase in parameters, requiring a large amount of data to train a well-performing model. In many cases, training data faces a scarcity issue. Therefore, this paper proposes MSCPT, which leverages Vision-Language models combined with pathological descriptions from Large language models to enhance the performance in few-shot scenarios.

\subsection{Vision-Language Models}
Vision-Language models (VLMs) are rapidly developing in various fields. During training, VLMs use contrastive learning to reduce distances between paired image-text pairs and increase distances between unpaired ones. The creators of CLIP~\cite{clip} collected over 400M image-text pairs from the internet and used contrastive learning to align them, resulting in compatibility across various tasks. Compared to natural images, gathering pairs of pathological images and corresponding descriptions is challenging. To address this issue, MI-Zero~\cite{mizero} first pre-trained image and text encoders using unpaired data and then aligns them in a common space using 33,480 pathological image-text pairs. Huang \textit{et al.} gathered over 208K pathological image-text pairs from Twitter and LAION~\cite{schuhmann2022laion} and developed PLIP~\cite{plip}. Lu \textit{et al.} trained Conch~\cite{conch} on over 1.17M pathological image-text pairs, and this model performs well on downstream tasks. Pre-trained VLMs have significant potential, but effective methods to leverage them for WSI-level tasks are lacking. In this paper, we propose using pathological visual descriptions as prior knowledge to unleash the potential of VLMs.

\subsection{Prompt Tuning in Vision-Language Models}
Prompt tuning has demonstrated remarkable efficiency and effectiveness, whether in text or multimodal domains~\cite{coop, cocoop, maple}. CLIP demonstrated remarkable zero-shot performance with hand-crafted prompts, but the results can vary significantly depending on the prompt used due to their sensitivity to changes. Therefore, CoOp~\cite{coop} and CoCoOp~\cite{cocoop} proposed that the model itself should determine the choice of prompts. Khattak \textit{et al.} argued that optimizing prompt tuning within a single branch is not ideal. They introduced MaPLe~\cite{maple} to enhance the alignment between visual and language representations. However, these innovative methods are highly applicable to natural images but do not consider the enormous size of WSIs and the crucial multi-scale and contextual information needed for WSI analysis. 
To the best of our knowledge, TOP~\cite{qu2024rise} was the first study to explore fine-tuning CLIP for FSWC tasks. Shortly thereafter, Shi \textit{et al.} introduced ViLa-MIL~\cite{shi2024vila}, which builds upon CLIP and enhances WSI classification by incorporating multi-scale language prior knowledge. These pioneering studies have significantly advanced the utility of VLMs and improved model performance in few-shot WSI classification scenarios.

However, both TOP and ViLa-MIL are CLIP-based and do not evaluate the performance of models on pathological VLMs. Moreover, due to the large number of patches in a WSI, these approaches primarily emphasize text while overlooking visual prompt tuning and the crucial contextual information within the WSI. Although TOP focuses on instance-level and batch-level prompt tuning, it does not incorporate multi-scale information. ViLa-MIL, on the other hand, incorporates multi-scale information but employs a late fusion strategy, which limits its ability to fully explore interactions across scales. Furthermore, while ViLa-MIL introduces category-specific priors, it is constrained by the context length of the CLIP text encoder (77 tokens), restricting the amount of prior knowledge it can utilize. Recent advances in pathology-specific VLMs~\cite{conch, seyfioglu2024quilt, sun2024pathgen, sun2024cpath} have significantly enhanced their ability to capture domain-specific pathological knowledge, motivating the development of MSCPT to better leverage detailed and effective prior knowledge.

\section{Method}
In this section, we introduce our few-shot weakly supervised WSI classification model, named \textbf{M}ulti-\textbf{S}cale and \textbf{C}ontext-focused \textbf{P}rompt \textbf{T}uning (MSCPT), as illustrated in Fig. ~\ref{figure2}. MSCPT utilizes a dual-path asymmetric structure as its foundation while conducting hierarchical prompt tuning on both textual and visual modalities.

\subsection{Problem Formulation}
Given a dataset $X = \left \{ X_1, X_2,..., X_N \right \}$ consisting of $N$ WSIs, each WSI is cropped into non-overlapping small patches, named instances. All instances belonging to the same WSI collectively form a bag. In weakly supervised WSI tasks, only the labels of bags are known. The labels of the bags $Y_i\in\left\{0,1\right\},\ i=\{1,2,...N\}$ and the label of each instance $\{y_{i,j},j=1,2,\ldots M_i\}$ have the following relationship:
\begin{equation}
\small
   Y_i=\left\{\begin{array}{cc}
      0,& \quad \text{if} \sum_j y_{i, j}=0, \\
      1,& \quad \text{else}.
   \end{array}\right.
   \label{eq1}
\end{equation}

\subsection{Review of CLIP and Patch Selection}
\subsubsection{Review of CLIP}
CLIP~\cite{clip} adopts a two-tower structure, including an image encoder and a text encoder. The image encoder $\mathcal{F}_{img}$ can be either a ResNet~\cite{resnet} or a ViT~\cite{vit}, which is used to transform images into visual embeddings. The text encoder $\mathcal{F}_{text}$ takes a series of words as input and outputs textual embeddings. During the training process, CLIP utilizes a contrastive loss to learn a joint embedding space for the two modalities. During inference, we assume $\boldsymbol{x}$ is the visual embedding, and $\left \{ \boldsymbol{w}_i \right \} _{i=1}^K$ is a series of textual embeddings generated by $\mathcal{F}_{text}$. Each $\boldsymbol{w}_i$ corresponds to a prompt (such as \textit{``an image of \{\textbf{class name}\}"}) embedding for a specific image category. Therefore, the predicted probabilities can be obtained by calculating the cosine similarity between $\boldsymbol{x}$ and $\boldsymbol{w}_i$:
\begin{eqnarray}
\small
p(y=i\mid  \boldsymbol{x}) = \frac{exp(cos(\boldsymbol{x},\boldsymbol{w}_i)/\tau)}{{\textstyle \sum_{j = 1}^{K}exp(cos(\boldsymbol{x},\boldsymbol{w}_j)/\tau)}},
\end{eqnarray}
where $\tau$ is the temperature coefficient, $cos(\cdot ,\cdot )$ represents the cosine similarity, and $K$ is the number of categories.

\subsubsection{Patch Selection}
Some studies have shown that simultaneously performing prompt tuning on the visual encoder while tuning the text encoder significantly improves the performance of VLMs across various downstream tasks~\cite{maple,metaprompt}. Due to the high resolution of WSIs, dividing them into non-overlapping patches will result in a large number of patches. Performing visual prompt tuning on all patches is computationally impractical due to the significant resource requirements. Fortunately, research has shown that only a few patches contain crucial information~\cite{clam}. By leveraging the robust zero-shot capabilities of VLMs to preliminarily identify patches most relevant to cancer analysis, the computational cost of visual prompt tuning can be significantly reduced.

Specifically, we utilize $\mathcal{F}_{img}$ to extract visual embeddings from patches while leveraging $\mathcal{F}_{text}$ to extract textual embeddings from the category prompts. For each class, a category prompt consists of two components: the class name (\textit{e.g.}, \textit{``squamous cell carcinoma"}) and the template (\textit{e.g.}, \textit{``a photomicrograph of \{\textbf{class name}\}"}), which together form the final prompt (\textit{e.g.}, \textit{``a photomicrograph of squamous cell carcinoma"}). To enhance the robustness of patch selection, we randomly generated 50 sets of prompts using various category names and templates. Following the method in~\cite{mizero}, we averaged the embeddings of these prompts. Subsequently, the similarities between patches and prompts are computed. Then, the top $n$ patches with the highest similarity scores are selected for each category. For a WSI $X_i$, we choose patches $\left \{ x^l_{i,j},j=1,2,...,n_l \right \}$ at low magnification. Due to our unique architecture, we solely perform patch selection and visual prompt tuning at low magnification.

\subsection{Multi-scale Visual Descriptions Construction}
In this part, we aim to generate pathological visual descriptions as \textbf{pathological language prior knowledge} to guide the hierarchical prompt tuning and instances aggregation. To reduce manual workload, large language models (LLMs) are employed to generate descriptions related to different diseases. That is, we enter the question \textit{``We are studying {\textbf{Cancer Category}}. Please list $\textbf{C}^l$ visual descriptions at $5\times$ magnification and $\textbf{C}^h$ visual descriptions at $20\times$ magnification observed in H\&E-stained histological images of {\textbf{Cancer Sub-category}}."} into the LLM. And then we can get the multi-scale visual description sets $T^{low} = \{ T^{low}_{k,c} \mid 0 \leq k \leq K, 0 \leq c \leq C^l \}$ and $T^{high} = \{ T^{high}_{k,c} \mid 0 \leq k \leq K, 0 \leq c \leq C^h \}$. $K$ represents the number of WSI categories, and $C^l$ and $C^h$ denote the counts of low-level and high-level descriptions, respectively.

\subsection{Multi-scale Hierarchical Prompt Tuning}
\label{prompt tuning}
Inspired by MetaPrompt~\cite{metaprompt}, a unique dual-path asymmetric framework is employed for multimodal hierarchical prompt tuning, as shown on the left of Fig. \ref{figure2}. Freezing two of the four encoders helps reduce the trainable parameters and alleviates overfitting in few-shot scenarios. Compared to previous works where their encoders process the same inputs, our method adopts a unique strategy: \textbf{the prompted and frozen encoders take entirely different inputs}. Considering the immense size of WSIs and the substantial computational and storage resource requirements for visual prompt tuning, we only conducted visual prompt tuning at the low level.

\begin{figure}[t!]
   \begin{center}
   % \fbox{\rule{0pt}{2in} \rule{0.9\linewidth}{0pt}}ƒ
   \includegraphics[width=1\linewidth]{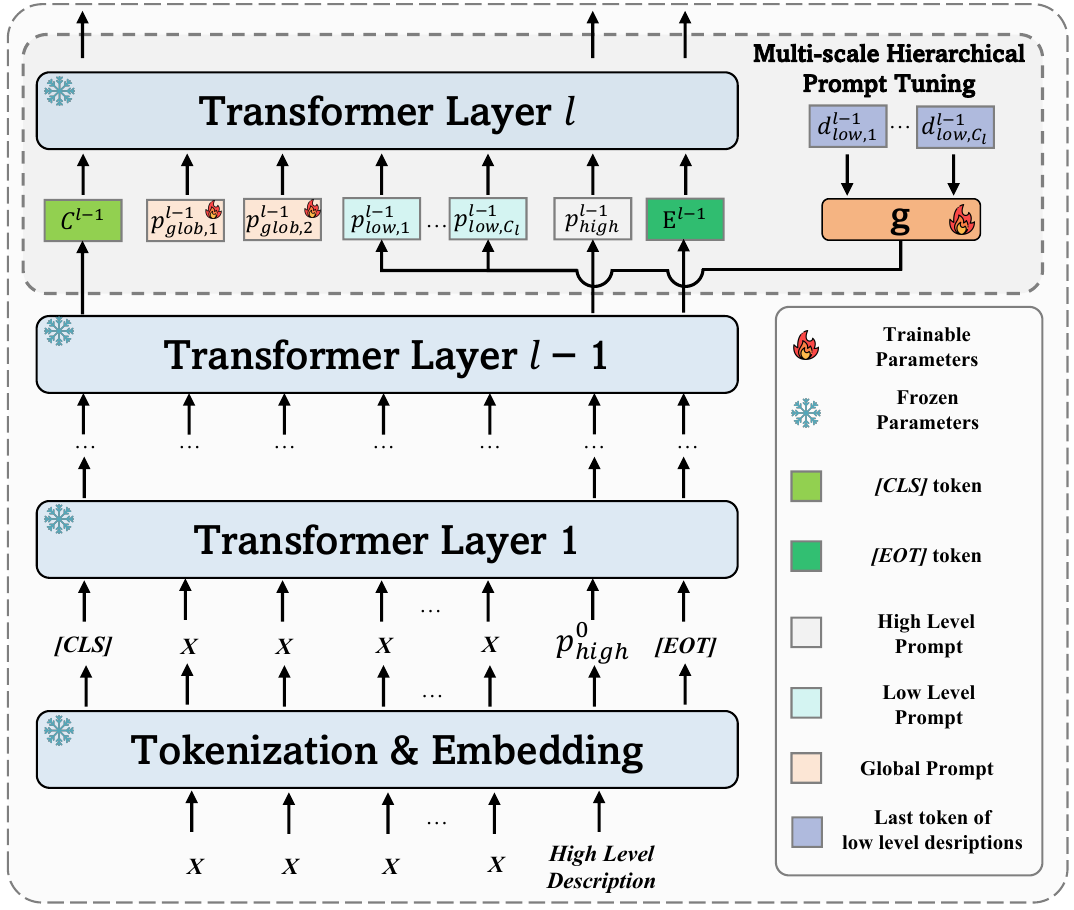}
   \end{center}
   \vspace{-0.5em}
   \caption{Details of the Prompted Hierarchical High-Level Text Encoder. The Multi-scale Hierarchical Prompt Turning (MHPT) module utilizes the transformer layer to integrate pathological visual descriptions from different scales.}
   \label{figure3}
   \vspace{-1.5em}
\end{figure}

Rather than modifying the visual prompts tuning method, our emphasis is placed on the text modality. Specifically, the low-level pathological visual descriptions $T^{low}$ are sent into the frozen low-level text encoder $\mathcal{F}^{low}_{text}$, while the high-level pathological visual descriptions $T^{high}$ are sent into the \textbf{prompted hierarchical high-level text encoder} $\mathcal{F}^{high}_{text}$. Although the low-level text encoder remains frozen, we propose a novel cross-level prompt generator that projects embeddings from the low-level encoder into the prompted hierarchical high-level text encoder during training. This design preserves the information encapsulated in low-level visual descriptions. Simultaneously, patches are also fed into the corresponding encoders. We wish to integrate different information contained in $T^{low}$ and $T^{high}$, which can enhance the multi-scale information processing capability of MSCPT. To achieve this purpose, we propose \textbf{M}ulti-scale \textbf{H}ierarchical \textbf{P}rompt \textbf{T}uning (MHPT) module. The core component of MHPT, the prompted hierarchical high-level text encoder, is illustrated in Fig. \ref{figure3}.

\subsubsection{Multi-scale Prompts Construction}
For each layer of $\mathcal{F}^{low}_{text}$, we introduce a learnable vector called global prompts $p_{glob}$ to learn and integrate information from high-level text prompts $p_{high}$ and low-level text prompts $p_{low}$. As an example, consider the construction of multi-scale prompts for a high-level pathological visual description $T^{high}_{k,c}$. After tokenization and embedding, $T^{high}_{k,c}$ is transformed into $p^{high}_0$. And then the low-level text prompts $p_{low}$ are obtained based on $T^{low}$. More specifically, a set of descriptions $T^{low}$ gets fed into the frozen $\mathcal{F}^{low}_{text}$, and the last token of each transformer layer gets extracted. These tokens are then fed into a prompt generator $g$, formulated as:
\begin{eqnarray}
\small
p_{low,i}^l=g\left ( d_{low}^l  \right ),
\end{eqnarray}
where $d_{low}^l$ is the last token of $T^{low}_{k,i}$ at the $l$-th layer, and generator $g$ is a basic multilayer perceptron to align vectors of different scales into a common embedding space. Then, these tokens get concatenated to obtain low-level text prompts $p_{low}^l$.
% $p_{low}^l=\left [p_{low,1}^l,..., p_{low,C^l}^l\right ] $.

\subsubsection{Hierarchical Prompt Tuning}
After obtaining the three prompts, to capture more complex associations between pathological visual descriptions at multi-scale, hierarchical prompt tuning is performed on $\mathcal{F}^{high}_{text}$, which can be expressed as:
\begin{align}
\begin{aligned}
\small
\left [ C^i,\_,\_,p^i_{high},E^i \right ] &=  T_i\left [C^{i-1}, p_{glob}^{i-1} ,p_{low}^{i-1},p_{high}^{i-1}, E^{i-1}\right ], \\
i &= 1,2,3,...,L,
\end{aligned}
\end{align}
where $C^i$ and $E^i$ represent the class token $\left [CLS\right ]$ and the last token $\left [EOT\right ]$ of the $i$-th transformer layer $T^i$, and $L$ signifies the number of transformer layers. Lastly, by projecting the last token of the last transformer layer through the textual projection head $TextProj$ into the joint embedding space, the final textual representation $z^{high}_{k,c}$ for $T^{high}_{k,c}$ is obtained:
\begin{align}
\small
z^{high}_{k,c} = Textproj(E^L).
\end{align}

\subsection{Image-text Similarity-based Graph Prompt Tuning}
Some studies have shown that the interactions between different areas of WSI and their structural information play a crucial role in cancer analysis~\cite{shao2023characterizing}. However, the current prompt tuning methods are unable to capture this information. To address this, we propose \textbf{I}mage-text \textbf{S}imilarity-based \textbf{G}raph \textbf{P}rompt \textbf{T}uning (ISGPT) module. More specifically, we deviate from conventional methods that utilize patch coordinates or patch feature similarity in constructing graph neural networks (GNNs)~\cite{patchgcn, h2gt}. Our innovative approach involves utilizing the similarity between patches and pathological visual descriptions as the foundation for developing GNNs. We treat the patches as nodes and aim to construct the adjacency matrix $\boldsymbol{A}$ by calculating the semantic similarity $\boldsymbol{S}$ between the patch embeddings and description embeddings. Specifically, after patches and descriptions have passed through the encoders from Section \ref{prompt tuning}, patch embeddings $\boldsymbol{P} \in \mathbb{R}^{M \times d}$ and description embeddings $\boldsymbol{Z} \in \mathbb{R}^{KC \times d}$ are obtained, respectively. The formula for semantic similarity $\boldsymbol{S} \in \mathbb{R}^{M \times KC}$ is:
\begin{eqnarray}
\small
s_{i,j} = \frac{exp(cos(\boldsymbol{P}_i,\boldsymbol{Z}_j)/\tau)}{{\textstyle \sum_{m = 1}^{K \times C}exp(cos(\boldsymbol{P}_i,\boldsymbol{Z}_m)/\tau)}},
\end{eqnarray}
where $\tau$ is the temperature coefficient, $cos(\cdot ,\cdot )$ represents the cosine similarity. $K$ represents the number of WSI categories and $d$ is the embedding dimensionality. $C$ and $M$ denote the number of pathological descriptions and patches at a given scale, respectively. Subsequently, the calculation formula for the adjacency matrix $\boldsymbol{A} \in \mathbb{R}^{M \times M}$ is written as:
\begin{eqnarray}
\small
a_{i,j} = \frac{exp(cos(\boldsymbol{S}_i,\boldsymbol{S}_j)/\tau)}{{\textstyle \sum_{m = 1}^{M}exp(cos(\boldsymbol{S}_i,\boldsymbol{S}_m)/\tau)}},
\end{eqnarray}
where $\boldsymbol{S}_i\in \mathbb{R}^{KC}$ represents the semantic similarity between $i$-th patch embeddings and all description embeddings (\textit{i.e.}, the $i$-th row of $\boldsymbol{S}$). We avoid constructing $\boldsymbol{A}$ based on patch coordinates or patch feature similarity, as this approach might overlook sparse but significant patches when focusing solely on Euclidean distance or feature similarity. Subsequent experimental results demonstrate the superior performance of our method for constructing $\boldsymbol{A}$. We choose a Graph Convolutional Network (GCN)~\cite{gcn} as the graph learning model. The GCN operation in the $l$-th layer is defined as:
\begin{align}
\small
\mathcal{F}_{GCN}\left ( \boldsymbol{A},\boldsymbol{H}^{\left( l \right)} \right )  &=  \sigma(\tilde{\boldsymbol{D}}^{-\frac{1}{2}}\tilde{\boldsymbol{A}}\tilde{\boldsymbol{D}}^{-\frac{1}{2}}\boldsymbol{H}^{\left ( l \right ) }\boldsymbol{W}^{\left ( l \right ) }).
\end{align}
Here $\tilde{\boldsymbol{A}} = \boldsymbol{A} + \boldsymbol{I}$, $\boldsymbol{I}$ is the identity matrix and $\sigma \left( \cdot \right)$ denotes an activation function. $\tilde{D}_{i,i}= {\textstyle \sum_{j}^{}\tilde{A}_{i,j}}$, $\boldsymbol{W}^{\left( l \right)}$ is layer-specific trainable weight matrix. $\boldsymbol{H}^{\left( l \right)} \in \mathbb{R} ^{M \times d}$ is the input embeddings of all nodes. Therefore, the patch embeddings after graph prompt tuning at both high and low scales are represented as:
\begin{align}
\small
\tilde{\boldsymbol{P}}^{high} & = \mathcal{F}^{high}_{GCN}(\boldsymbol{A}^{high},\boldsymbol{P}^{high}), \\
\tilde{\boldsymbol{P}}^{low} & = \mathcal{F}^{low}_{GCN}(\boldsymbol{A}^{low},\boldsymbol{P}^{low}).
\end{align}

\subsection{Non-Parametric Cross-Guided Pooling}
Motivated by the powerful zero-shot capability of pre-trained VLMs, the possibility of employing a similar non-parametric approach for instance aggregation was pondered. We propose \textbf{N}on-\textbf{P}arametric \textbf{C}ross-\textbf{G}uided \textbf{P}ooling (NPCGP) to aggregate instances into bag features. In NPCGP, we compute semantic similarities between the patch embeddings $\tilde{\textbf{P}}$ post graph tuning and pathological visual description embeddings $\boldsymbol{Z}$ at both the same and across scales. The reason for calculating similarities both within the same and across scales is our concern that the pathological visual descriptions provided by LLM may contain scale-related inaccuracies. Hence, this procedure serves to bolster the robustness of feature aggregation strategies. Lastly, the bag-level unnormalized probability distribution $Logits$ is obtained through the Top-K max-pooling operator $h_{topK}$:
\begin{eqnarray}
\small
\begin{aligned}
Logits^{high} &= h_{\text{topK}}\left( \tilde{\mathbf{P}}^{high} \cdot {\mathbf{Z}^{high}}^\mathrm{T} \right) \\
&\quad + h_{\text{topK}}\left( \tilde{\mathbf{P}}^{high} \cdot {\mathbf{Z}^{low}}^\mathrm{T} \right),
\end{aligned} \\
\begin{aligned}
Logits^{low} &= h_{\text{topK}}\left( \tilde{\mathbf{P}}^{low} \cdot {\mathbf{Z}^{low}}^\mathrm{T} \right) \\
&\quad + h_{\text{topK}}\left( \tilde{\mathbf{P}}^{low} \cdot {\mathbf{Z}^{high}}^\mathrm{T} \right) ,
\end{aligned} \\
\begin{aligned}
 Logits^{overall} &=\frac{1}{2} \left (Logits^{high} + Logits^{low}\right ).
\end{aligned}
\end{eqnarray}
Following~\cite{metaprompt}, we use cross-entropy loss to optimize the three distributions $Logits^{overall}$, $Logits^{high}$, and $Logits^{low}$, but only $Logits^{overall}$ was used during model inference.

\begin{table*}[t]
\centering
\caption{Results on five datasets, three downstream tasks and three VLMs under the 16-shot setting. The highest performance is in bold, the second-best underlined, with mean and standard deviation reported across five random seeds.}
\resizebox{\textwidth}{!}{
\begin{tblr}{
  colsep=1.5pt,
  rowsep = 0.05cm,
  column{3} = {c},
  column{4} = {c},
  column{5} = {c},
  column{6} = {c},
  column{7} = {c},
  column{8} = {c},
  column{9} = {c},
  column{10} = {c},
  column{11} = {c},
  column{12} = {c},
  column{13} = {c},
  column{14} = {c},
  column{15} = {c},
  column{16} = {c},
  column{17} = {c},
  column{18} = {c},
  column{19} = {c},
  column{20} = {c},
  column{21} = {c},
  column{22} = {c},
  column{23} = {c},
  row{16} = {bg=lightgray},
  row{30} = {bg=lightgray},
  row{44} = {bg=lightgray},
  cell{1}{1} = {c=2,r=2}{c},
  cell{1}{3} = {r=2}{},
  cell{1}{4} = {c=3}{},
  cell{1}{8} = {c=3}{},
  cell{1}{12} = {c=3}{},
  cell{1}{16} = {c=3}{},
  cell{1}{20} = {c=3}{},
  cell{3}{1} = {r=14}{c, bg=lightgray},
  cell{17}{1} = {r=14}{c, bg=lightgray},
  cell{31}{1} = {r=14}{c, bg=lightgray},
  hline{1,17,31,45} = {-}{0.08em},
  hline{2} = {4-6,8-10,12-14,16-18,20-22}{},
  hline{3} = {-}{},
}
\textbf{Methods}       &               & \textbf{\makecell{Train. \\ Param}}   & {\textbf{TCGA-NSCLC} \\ (Subtyping, 2 classes)}  & &        & & {\textbf{TCGA-RCC} \\ (Subtyping, 3 classes)}  &            &      &  & {\textbf{UBC-OCEAN} \\ (Subtyping, 5 classes)}     & &        & & {\textbf{TCGA-BRCA} \\ (Recurrence Prediction, 2 classes)}     & &        & & {\textbf{PANDA} \\ (Grading, 3 classes)}         \\
                                                                       &                      &        &   AUC         & F1          & ACC        &  & AUC         & F1         & ACC         &  & AUC        & F1         & ACC  &  & AUC        & F1         & ACC &  & AUC        & F1         & ACC      \\
\begin{sideways}\makecell{\large \textbf{CLIP}\\  Pretrained on 400M Natural Image Pairs }\end{sideways}              
                                                                       & Max-pooling          & 197K     & 63.80$_{\pm6.84}$  & 60.40$_{\pm4.76}$  & 60.70$_{\pm4.75}$ &  & 84.51$_{\pm3.21}$ & 65.83$_{\pm2.72}$ & 69.26$_{\pm2.33}$ &  & 87.76$_{\pm1.45}$  & 62.36$_{\pm5.63}$ & 63.58$_{\pm4.05}$ 
                                                                       & &
                                                                       59.74$_{\pm1.88}$ &
                                                                       57.03$_{\pm1.71}$ &
                                                                       63.85$_{\pm2.06}$ 
                                                                       & &
                                                                       72.12$_{\pm1.71}$ &
                                                                       51.68$_{\pm3.55}$ &
                                                                       56.48$_{\pm1.50}$
                                                                       \\
                                                                       & Mean-pooling         & 197K     & 69.53$_{\pm4.74}$  & 63.76$_{\pm5.77}$  & 64.69$_{\pm4.31}$ &  & 93.31$_{\pm0.66}$ & 78.64$_{\pm0.74}$ & 81.29$_{\pm1.08}$ &  & 85.18$_{\pm1.70}$  & 59.59$_{\pm4.02}$ & 62.16$_{\pm3.75}$ & &64.19$_{\pm3.40}$  & 58.08$_{\pm3.38}$ & 61.01$_{\pm4.19}$ 
                                                                       & &
                                                                       77.85$_{\pm1.32}$ &
                                                                       58.86$_{\pm1.69}$ &
                                                                       62.75$_{\pm1.96}$
                                                                       \\
                                                                       & ABMIL\cite{abmil}   & 461K     & 66.95$_{\pm4.31}$  & 62.60$_{\pm3.75}$  & 62.96$_{\pm3.65}$ &  & 93.41$_{\pm1.41}$ & 79.80$_{\pm1.56}$ & 82.47$_{\pm1.46}$ &  & 88.03$_{\pm0.98}$  & 65.33$_{\pm1.57}$ & 65.98$_{\pm2.02}$
                                                                       & &
                                                                       62.51$_{\pm4.10}$ &
                                                                       57.44$_{\pm3.51}$ &
                                                                       63.42$_{\pm3.00}$ 
                                                                       & &
                                                                       73.89$_{\pm2.40}$ &
                                                                       51.07$_{\pm2.87}$ &
                                                                       57.93$_{\pm2.99}$ &
                                                                       \\
                                                                       & CLAM-SB\cite{clam}  & 660K     & 67.49$_{\pm5.94}$  & 62.86$_{\pm4.19}$  & 63.51$_{\pm4.19}$ &  & 93.85$_{\pm1.52}$ & 79.87$_{\pm3.17}$ & 83.21$_{\pm2.67}$ &  & 88.10$_{\pm1.55}$ & 66.11$_{\pm2.88}$ & 66.76$_{\pm2.97}$  
                                                                       & &
                                                                       \uline{65.79$_{\pm2.89}$} &
                                                                       59.94$_{\pm3.09}$ &
                                                                       63.53$_{\pm4.64}$
                                                                       & &
                                                                       76.66$_{\pm1.53}$ &
                                                                       56.92$_{\pm2.87}$ &
                                                                       58.76$_{\pm3.71}$ &
                                                                       \\
                                                                       & CLAM-MB\cite{clam}  & 660K     & 69.65$_{\pm3.61}$  & 64.52$_{\pm3.22}$  & 65.14$_{\pm2.69}$ &  & 93.59$_{\pm1.16}$ & 78.72$_{\pm2.18}$ & 81.03$_{\pm2.06}$ & & 87.09$_{\pm0.73}$  & 63.54$_{\pm2.20}$ & 65.69$_{\pm1.64}$ 
                                                                       & &
                                                                       64.74$_{\pm3.37}$ &
                                                                       59.59$_{\pm3.23}$ &
                                                                       64.12$_{\pm4.83}$ 
                                                                       & &
                                                                       74.68$_{\pm2.43}$ &
                                                                       55.40$_{\pm3.06}$ &
                                                                       58.08$_{\pm3.61}$ 
                                                                       \\
                                                                       & TransMIL\cite{shao2021transmil}    & 2.54M   & 64.82$_{\pm8.01}$  & 59.17$_{\pm10.87}$ & 62.00$_{\pm5.18}$ &  & 94.17$_{\pm1.23}$ & 79.63$_{\pm1.52}$ & 81.86$_{\pm1.41}$ &  & 83.49$_{\pm1.77}$  & 54.36$_{\pm4.74}$ & 55.59$_{\pm5.88}$ 
                                                                       & &
                                                                       60.85$_{\pm4.37}$ &
                                                                       55.97$_{\pm3.21}$ &
                                                                       60.93$_{\pm3.39}$ 
                                                                       & &
                                                                       71.95$_{\pm1.86}$ &
                                                                       53.68$_{\pm3.00}$ &
                                                                       57.43$_{\pm3.23}$ &
                                                                       \\
                                                                       & DSMIL\cite{dsmil}   & 462K   & 66.00$_{\pm9.23}$  & 63.87$_{\pm7.00}$  & 64.11$_{\pm6.65}$ &  & 91.53$_{\pm5.17}$ & 78.38$_{\pm6.56}$ & 80.69$_{\pm6.47}$ &  & 87.55$_{\pm0.55}$ & 64.84$_{\pm1.80}$ & 66.23$_{\pm1.89}$
                                                                       & &
                                                                       60.42$_{\pm3.07}$ &
                                                                       57.82$_{\pm2.58}$ &
                                                                       63.44$_{\pm2.52}$ 
                                                                       & &
                                                                       72.14$_{\pm3.36}$ &
                                                                       51.79$_{\pm6.61}$ &
                                                                       58.69$_{\pm4.25}$ &
                                                                       \\
                                                                       & RRT-MIL\cite{RRTMIL}& 2.63M  & 66.47$_{\pm6.73}$  & 62.10$_{\pm6.17}$  & 63.20$_{\pm5.24}$ &  & 93.89$_{\pm1.91}$ & \uline{81.04$_{\pm2.11}$} & \uline{83.30$_{\pm2.24}$} &  & 88.02$_{\pm1.10}$  & 62.40$_{\pm2.79}$ & 64.22$_{\pm3.02}$
                                                                       & &
                                                                       62.38$_{\pm4.82}$ &
                                                                       57.80$_{\pm5.76}$ &
                                                                       61.21$_{\pm4.60}$ 
                                                                       & &
                                                                       74.03$_{\pm5.59}$ &
                                                                       53.78$_{\pm7.21}$ &
                                                                       60.27$_{\pm5.31}$ &
                                                                       \\
                                                                       & CoOp\cite{coop}     & 337K   & 69.06$_{\pm4.06}$  & 63.87$_{\pm3.77}$  & 64.27$_{\pm3.55}$  &  & 94.18$_{\pm1.72}$ & 79.88$_{\pm2.40}$ & 82.15$_{\pm1.96}$ &  & 88.65$_{\pm0.92}$  & 66.82$_{\pm2.53}$ & 68.04$_{\pm2.11}$  
                                                                       & &
                                                                       64.16$_{\pm3.37}$ &
                                                                       58.69$_{\pm3.08}$ &
                                                                       62.89$_{\pm2.67}$ 
                                                                       & &
                                                                       78.71$_{\pm3.03}$ &
                                                                       60.38$_{\pm4.01}$ &
                                                                       64.37$_{\pm3.31}$ 
                                                                       \\
                                                                       & CoCoOp\cite{cocoop} & 370K   & 64.37$_{\pm2.28}$  & 60.95$_{\pm1.55}$  & 61.37$_{\pm1.36}$ &  & 85.68$_{\pm2.66}$ & 67.72$_{\pm3.49}$ & 71.00$_{\pm2.90}$ &  & 88.57$_{\pm1.07}$  & 66.93$_{\pm1.69}$ & 67.70$_{\pm1.70}$ 
                                                                       & &
                                                                       63.62$_{\pm3.79}$ &
                                                                       58.76$_{\pm2.39}$ &
                                                                       62.78$_{\pm3.55}$ 
                                                                       & &
                                                                       77.01$_{\pm1.81}$ &
                                                                       55.02$_{\pm5.33}$ &
                                                                       60.56$_{\pm2.48}$ &
                                                                       \\
                                                                       & Metaprompt\cite{metaprompt}& 688K   & \uline{75.94$_{\pm3.01}$}  & \uline{70.35$_{\pm3.09}$}  & \uline{70.41$_{\pm3.09}$}     &  & \uline{94.18$_{\pm1.56}$} & 80.03$_{\pm2.06}$ & 82.52$_{\pm2.15}$ & & \uline{88.79$_{\pm0.76}$}  & \uline{67.43$_{\pm1.82}$} & \uline{68.56$_{\pm2.22}$}   
                                                                       & &
                                                                       65.78$_{\pm2.73}$ &
                                                                    \uline{60.21$_{\pm3.49}$} &
                                                                    \uline{64.22$_{\pm2.21}$} 
                                                                       & &
                                                                       78.87$_{\pm2.51}$ &
                                                                       59.90$_{\pm3.81}$ &
                                                                       62.71$_{\pm3.71}$ 
                                                                       \\
                                                                       & TOP\cite{qu2024rise}                  & 2.11M  & 73.56$_{\pm3.14}$       & 68.19$_{\pm1.22}$       & 68.77$_{\pm2.53}$     &  & 93.56$_{\pm1.22}$      & 79.66$_{\pm1.97}$      & 80.79$_{\pm1.05}$  &  & 87.93$_{\pm 0.88}$       & 66.74$_{\pm 2.44}$      & 67.15$_{\pm 1.82}$ 
                                                                       & &
                                                                       63.18$_{\pm2.44}$ &
                                                                       57.96$_{\pm4.14}$ &
                                                                       63.91$_{\pm3.12}$ 
                                                                       & &
                                                                       78.55$_{\pm2.34}$ &
                                                                       58.83$_{\pm3.67}$ &
                                                                       61.80$_{\pm3.51}$
                                                                       \\
                                                                       & ViLa-MIL\cite{shi2024vila}             & 2.77M  & 74.85$_{\pm7.62}$  & 68.74$_{\pm5.86}$  & 68.87$_{\pm5.97}$ &  & 93.34$_{\pm1.49}$ & 79.40$_{\pm1.13}$ & 81.81$_{\pm0.92}$ &  &  87.19$_{\pm1.03}$  &
                                                                64.43$_{\pm1.84}$ & 65.49$_{\pm1.36}$   
                                                                       & &
                                                                       64.00$_{\pm2.18}$ &
                                                                       57.59$_{\pm3.41}$ &
                                                                       63.67$_{\pm1.98}$ 
                                                                       & &
                                                              \uline{79.63$_{\pm1.86}$ }&
                                                                \uline{61.11$_{\pm3.25}$} &
                                                                \uline{65.57$_{\pm2.49}$} &
                                                                       \\
                                                                      & \textbf{MSCPT(ours)} & 1.35M  & \textbf{78.67$_{\pm3.93}$}  & \textbf{72.47$_{\pm3.13}$}  & \textbf{72.67$_{\pm2.96}$}  &  & \textbf{95.04$_{\pm1.31}$} & \textbf{83.78$_{\pm2.19}$} & \textbf{85.62$_{\pm2.14}$} &  & \textbf{89.99$_{\pm0.93}$}  & \textbf{68.14$_{\pm3.65}$} &\textbf{70.10$_{\pm3.61}$}   
                                                                      & &
                                                                \textbf{67.41$_{\pm5.35}$} &
                                                                \textbf{61.88$_{\pm3.48}$} &
                                                                \textbf{66.00$_{\pm2.70}$} 
                                                                      & &
                                                                \textbf{81.81$_{\pm2.01}$} &
                                                                \textbf{63.06$_{\pm2.93}$} &
                                                                \textbf{68.02$_{\pm2.96}$} &
                                                                      \\
\begin{sideways}\makecell{\large \textbf{PLIP}\\ Pretrained on 208K Pathology Image Pairs }\end{sideways}
                                                                       & Max-pooling          & 197K   & 71.78$_{\pm4.13}$   & 66.40$_{\pm3.51}$  & 66.66$_{\pm3.42}$ &  & 95.18$_{\pm0.63}$ & 81.63$_{\pm0.92}$ & 84.30$_{\pm1.30}$ &  & 91.57$_{\pm1.35}$  & 71.06$_{\pm3.30}$ & 71.09$_{\pm3.39}$   
                                                                       & &
                                                                       65.83$_{\pm2.66}$ &
                                                                       59.92$_{\pm1.62}$ &
                                                                       66.12$_{\pm3.57}$ 
                                                                       & &
                                                                       78.72$_{\pm3.05}$ &
                                                                       59.24$_{\pm3.22}$ &
                                                                       62.86$_{\pm3.13}$ &
                                                                       \\
                                                                       & Mean-pooling         & 197K   & 70.55$_{\pm6.64}$  & 65.32$_{\pm5.60}$  & 65.50$_{\pm5.55}$ &  & 94.75$_{\pm0.51}$ & 82.22$_{\pm0.67}$ & 85.24$_{\pm0.80}$ &  & 87.98$_{\pm1.06}$  & 62.61$_{\pm1.64}$ & 65.05$_{\pm1.82}$  
                                                                       & &
                                                                       67.20$_{\pm3.34}$ &
                                                                       61.64$_{\pm2.86}$ &
                                                                       62.33$_{\pm3.87}$ 
                                                                       & &
                                                                       79.21$_{\pm1.74}$ &
                                                                       60.94$_{\pm0.80}$ &
                                                                       64.28$_{\pm1.55}$ &
                                                                       \\
                                                                       & ABMIL\cite{abmil}                & 461K   & 78.54$_{\pm4.29}$  & 72.06$_{\pm3.79}$  & 72.12$_{\pm3.78}$ &  & 96.51$_{\pm0.63}$ & \uline{85.66$_{\pm1.97}$} & \uline{87.94$_{\pm1.92}$} &  & 90.93$_{\pm0.33}$  & 70.82$_{\pm1.52}$ & 70.82$_{\pm1.98}$ 
                                                                       & &
                                                                       67.89$_{\pm3.17}$ &
                                                                       63.24$_{\pm1.71}$ &
                                                                       66.94$_{\pm3.75}$ 
                                                                       & &
                                                                       81.19$_{\pm4.25}$ &
                                                                       64.04$_{\pm4.56}$ &
                                                                       66.67$_{\pm4.08}$ &
                                                                       \\
                                                                       & CLAM-SB\cite{clam}              & 660K   & 80.56$_{\pm4.57}$  & 73.15$_{\pm4.05}$  & 73.27$_{\pm3.97}$ &  & 96.41$_{\pm0.36}$ & 84.71$_{\pm1.60}$ & 87.25$_{\pm1.34}$ &  & 91.54$_{\pm1.16}$  & 71.39$_{\pm3.13}$ & 71.65$_{\pm2.86}$  
                                                                       & &
                                                                    \uline{69.54$_{\pm3.37}$} &
                                                                    \uline{64.81$_{\pm2.40}$} &
                                                                    \uline{68.51$_{\pm3.06}$}
                                                                       & &
                                                                    \uline{82.84$_{\pm3.59}$} &
                                                                    \uline{65.39$_{\pm3.86}$} &
                                                                     67.76$_{\pm3.22}$ &
                                                                       \\
                                                                       & CLAM-MB\cite{clam}              & 660K   & 80.68$_{\pm3.63}$  & 73.15$_{\pm3.00}$  & 73.32$_{\pm2.83}$ &  & \uline{96.58$_{\pm0.59}$} & 85.20$_{\pm0.83}$ & 87.85$_{\pm0.79}$ &  & \uline{91.66$_{\pm0.95}$}  & \uline{71.55$_{\pm3.18}$} & 71.62$_{\pm3.19}$  
                                                                       & &
                                                                       68.84$_{\pm4.10}$ &
                                                                       64.31$_{\pm1.09}$ &
                                                                       67.97$_{\pm3.00}$
                                                                       & &
                                                                       82.07$_{\pm3.45}$ &
                                                                       64.81$_{\pm3.96}$ &
                                                                       67.53$_{\pm4.00}$ &
                                                                       \\
                                                                       & TransMIL\cite{shao2021transmil}             & 2.54M  & 73.40$_{\pm10.33}$  & 66.92$_{\pm7.94}$  & 67.21$_{\pm7.63}$ &  & 96.35$_{\pm0.54}$ & 83.70$_{\pm0.80}$ & 86.33$_{\pm0.46}$ &  & 91.64$_{\pm1.76}$  & 66.99$_{\pm2.91}$ & 69.51$_{\pm2.31}$
                                                                       & &
                                                                       67.10$_{\pm5.76}$ &
                                                                       60.28$_{\pm5.45}$ &
                                                                       65.60$_{\pm4.73}$ 
                                                                       & &
                                                                       79.86$_{\pm2.18}$ &
                                                                       61.87$_{\pm1.53}$ &
                                                                       65.25$_{\pm1.81}$ &
                                                                       
                                                                       \\
                                                                       & DSMIL\cite{dsmil}                & 462K   & 77.75$_{\pm7.22}$  & 72.84$_{\pm6.31}$  & 73.08$_{\pm6.00}$ &  & 93.01$_{\pm6.05}$ & 79.58$_{\pm9.16}$ & 82.87$_{\pm7.32}$ &  & 90.40$_{\pm1.24}$  & 68.54$_{\pm5.13}$ & 70.05$_{\pm3.44}$  
                                                                       & &
                                                                       63.59$_{\pm3.53}$ &
                                                                       61.22$_{\pm3.38}$ &
                                                                       65.16$_{\pm3.68}$ 
                                                                       & &
                                                                       72.14$_{\pm3.36}$ &
                                                                       51.79$_{\pm6.61}$ &
                                                                       58.69$_{\pm4.25}$ &
                                                                       \\
                                                                       & RRT-MIL\cite{RRTMIL}              & 2.63M  & 76.30$_{\pm10.01}$ & 70.86$_{\pm7.47}$  & 71.01$_{\pm7.44}$ &  & 96.09$_{\pm1.06}$ & 83.94$_{\pm2.05}$ & 86.56$_{\pm2.28}$ &  & 91.65$_{\pm1.13}$  & 68.37$_{\pm2.36}$ & 69.80$_{\pm2.44}$  
                                                                       & &
                                                                       66.20$_{\pm5.57}$ &
                                                                       62.09$_{\pm4.99}$ &
                                                                       65.19$_{\pm5.09}$ 
                                                                       & &
                                                                       82.05$_{\pm3.24}$ &
                                                                       63.43$_{\pm3.13}$ &
                                                                       66.22$_{\pm3.20}$ &
                                                                       \\
                                                                       & CoOp\cite{coop}                 & 337K   & 77.92$_{\pm5.48}$  & 71.58$_{\pm4.74}$  & 71.63$_{\pm4.75}$ &  & 95.76$_{\pm0.80}$ & 83.23$_{\pm2.07}$ & 85.90$_{\pm1.63}$ &  &  91.53$_{\pm0.73}$  & 71.21$_{\pm3.51}$ & 71.60$_{\pm3.93}$  
                                                                       & &
                                                                       68.47$_{\pm4.96}$ &
                                                                       64.26$_{\pm2.45}$ &
                                                                       66.71$_{\pm3.76}$ 
                                                                       & &
                                                                       81.74$_{\pm2.86}$ &
                                                                       64.81$_{\pm2.90}$ &
                                                                \uline{67.80$_{\pm3.11}$} &
                                                                       \\
                                                                       & CoCoOp\cite{cocoop}               & 370K   & 72.62$_{\pm8.45}$  & 66.63$_{\pm5.83}$  & 66.97$_{\pm5.85}$ &  & 95.81$_{\pm0.42}$ & 83.18$_{\pm1.35}$ & 86.02$_{\pm1.03}$ &  & 91.65$_{\pm0.70}$  & 71.48$_{\pm2.25}$ & \uline{72.01$_{\pm2.15}$}   
                                                                       & &
                                                                       67.49$_{\pm4.76}$ &
                                                                       64.57$_{\pm2.19}$ &
                                                                       67.24$_{\pm3.66}$ 
                                                                       & &
                                                                       81.36$_{\pm2.23}$ &
                                                                       62.42$_{\pm2.08}$ &
                                                                       65.59$_{\pm2.93}$
                                                                       \\
                                                                       & Metaprompt\cite{metaprompt}           & 688K   & 78.31$_{\pm5.66}$  & 72.03$_{\pm4.60}$  & 71.86$_{\pm4.61}$ &  & 95.75$_{\pm0.48}$ & 83.52$_{\pm1.46}$ & 86.62$_{\pm1.43}$ &  & 91.44$_{\pm0.30}$  & 70.90$_{\pm1.95}$ & 71.12$_{\pm1.85}$ 
                                                                       & &
                                                                       68.77$_{\pm4.54}$ &
                                                                       63.97$_{\pm3.38}$ &
                                                                       67.85$_{\pm3.18}$ 
                                                                       & &
                                                                       80.94$_{\pm2.32}$ &
                                                                       62.19$_{\pm3.48}$ &
                                                                       65.70$_{\pm2.96}$
                                                                       \\
                                                                       & TOP\cite{qu2024rise}                  & 2.11M  & 78.91$_{\pm3.79}$       & 72.33$_{\pm4.89}$       & 72.91$_{\pm4.61}$   &  & 95.06$_{\pm0.51}$      & 82.86$_{\pm1.35}$      & 86.14$_{\pm0.98}$     &  & 91.18$_{\pm 0.76}$       & 71.16$_{\pm 1.24}$      & 71.67$_{\pm 2.08}$      & &
                                                                       68.52$_{\pm4.07}$ &
                                                                       64.73$_{\pm2.65}$ &
                                                                       67.70$_{\pm4.01}$ 
                                                                       & &
                                                                       81.03$_{\pm2.49}$ &
                                                                       62.87$_{\pm3.57}$ &
                                                                       66.46$_{\pm3.15}$
                                                                       \\
                                                                       & ViLa-MIL\cite{shi2024vila}             & 2.77M  & \uline{80.98$_{\pm2.52}$}  & \uline{73.81$_{\pm3.64}$}  & \uline{73.94$_{\pm3.56}$} &  & 95.72$_{\pm0.60}$ & 83.85$_{\pm1.10}$ & 86.53$_{\pm1.03}$ &  & 91.00$_{\pm0.39}$  & 70.42$_{\pm1.02}$ & 70.98$_{\pm1.17}$ 
                                                                       & &
                                                                       69.24$_{\pm2.60}$ &
                                                                       63.80$_{\pm1.67}$ &
                                                                       67.92$_{\pm1.69}$  
                                                                       & &
                                                                       82.17$_{\pm2.73}$ &
                                                                       64.35$_{\pm3.15}$ &
                                                                       66.70$_{\pm3.56}$
                                                                       \\
                                                   & \textbf{MSCPT(ours)} & 1.35M  & \textbf{84.29$_{\pm3.97}$}  & \textbf{76.39$_{\pm5.69}$}  & \textbf{76.54$_{\pm5.49}$} &  & \textbf{96.94$_{\pm0.36}$} & \textbf{87.01$_{\pm1.51}$} & \textbf{89.28$_{\pm1.22}$} &  & \textbf{92.59$_{\pm1.92}$}  & \textbf{73.10$_{\pm2.71}$} & \textbf{73.65$_{\pm3.67}$}     
                                                   & &
                                                   \textbf{71.49$_{\pm2.68}$} &
                                                   \textbf{66.99$_{\pm3.78}$} &
                                                   \textbf{70.19$_{\pm3.11}$}
                                                   & &
                                                   \textbf{84.90$_{\pm3.85}$} &
                                                   \textbf{68.39$_{\pm3.96}$} &
                                                   \textbf{70.55$_{\pm4.50}$} &
                                                   \\
\begin{sideways}\makecell{\large \textbf{CONCH}\\ Pretrained on 1.17M Pathology Image Pairs}\end{sideways}
                                                                       & Max-pooling          & 131K   & 94.68$_{\pm1.91}$   & 87.97$_{\pm1.85}$  & 87.98$_{\pm1.85}$ &  & 97.42$_{\pm0.25}$ & 90.20$_{\pm0.91}$ & 91.35$_{\pm0.65}$ &  & 95.18$_{\pm0.56}$  & 80.78$_{\pm1.77}$ & 80.99$_{\pm1.87}$  
                                                                       & &
                                                                       76.00$_{\pm2.11}$ &
                                                                       67.18$_{\pm2.93}$ &
                                                                       71.38$_{\pm3.17}$ 
                                                                       & &
                                                                       84.42$_{\pm2.57}$ &
                                                                       65.90$_{\pm3.02}$ &
                                                                       69.85$_{\pm3.62}$
                                                                       \\
                                                                       & Mean-pooling         & 131K   & 90.09$_{\pm2.74}$  & 82.11$_{\pm3.98}$  & 82.12$_{\pm3.97}$ &  & 97.28$_{\pm0.23}$ & 88.05$_{\pm0.28}$ & 89.86$_{\pm0.59}$ &  & 94.59$_{\pm0.47}$  & 76.98$_{\pm1.46}$ & 78.48$_{\pm1.19}$   
                                                                       & &
                                                                       77.71$_{\pm3.05}$ &
                                                                       68.60$_{\pm2.72}$ &
                                                                       72.13$_{\pm3.05}$ 
                                                                       & &
                                                                       82.10$_{\pm1.20}$ &
                                                                       64.26$_{\pm0.74}$ &
                                                                       68.75$_{\pm1.29}$
                                                                       \\
                                                                       & ABMIL\cite{abmil}                & 329K   & 94.69$_{\pm2.17}$  & 87.80$_{\pm2.61}$  & 87.84$_{\pm2.57}$ &  & \uline{98.41$_{\pm0.50}$} & 91.41$_{\pm1.15}$ & \uline{92.69$_{\pm1.05}$} &  & 96.02$_{\pm0.36}$  & 81.24$_{\pm1.09}$ & 82.35$_{\pm1.09}$  
                                                                       & &
                                                                    \uline{79.40$_{\pm1.17}$} &
                                                                    \uline{70.19$_{\pm2.26}$} &
                                                                    \uline{74.95$_{\pm2.97}$} 
                                                                       & &
                                                                       86.98$_{\pm1.30}$ &
                                                                       69.99$_{\pm1.57}$ &
                                                                       72.97$_{\pm1.40}$ 
                                                                       \\
                                                                       & CLAM-SB\cite{clam}              & 528K   & \textbf{96.64$_{\pm1.44}$}  & \uline{88.92$_{\pm2.27}$}  & \uline{88.93$_{\pm2.27}$} &  & 98.02$_{\pm0.37}$ & 91.07$_{\pm1.34}$ & 91.85$_{\pm1.08}$ &  & \textbf{96.23$_{\pm0.35}$}  & \uline{82.19$_{\pm1.41}$} & \uline{82.74$_{\pm0.94}$}   
                                                                       & &
                                                                       79.34$_{\pm2.44}$ &
                                                                       70.02$_{\pm2.88}$ &
                                                                       73.70$_{\pm3.10}$ 
                                                                       & &
                                                                    \uline{87.95$_{\pm2.02}$} &
                                                                    \uline{71.75$_{\pm1.68}$} &
                                                                    \uline{74.68$_{\pm2.32}$} 
                                                                       \\
                                                                       & CLAM-MB\cite{clam}              & 528K   & 95.85$_{\pm1.28}$  & 88.78$_{\pm1.45}$  & 88.78$_{\pm1.44}$ &  & 97.88$_{\pm0.43}$ & 90.74$_{\pm1.35}$ & 91.59$_{\pm1.11}$ &  & 95.54$_{\pm0.23}$  & 81.21$_{\pm1.40}$ & 82.07$_{\pm0.94}$  
                                                                       & &
                                                                       78.93$_{\pm3.05}$ &
                                                                       70.06$_{\pm2.57}$ &
                                                                       74.45$_{\pm1.97}$ 
                                                                       & &
                                                                       87.22$_{\pm1.43}$ &
                                                                       71.08$_{\pm1.93}$ &
                                                                       73.93$_{\pm2.16}$
                                                                       \\
                                                                       & TransMIL\cite{shao2021transmil}             & 2.40M  & 93.29$_{\pm3.46}$  & 86.42$_{\pm4.03}$  & 86.44$_{\pm4.05}$ &  & 98.29$_{\pm0.25}$ & 90.56$_{\pm0.71}$ & 92.12$_{\pm0.75}$ &  & 95.67$_{\pm0.60}$  & 78.78$_{\pm0.94}$ & 80.20$_{\pm1.75}$  
                                                                       & &
                                                                       77.94$_{\pm3.54}$ &
                                                                       67.94$_{\pm4.32}$ &
                                                                       70.85$_{\pm4.50}$ 
                                                                       & &
                                                                       86.09$_{\pm1.55}$ &
                                                                       69.83$_{\pm2.54}$ &
                                                                       72.97$_{\pm2.98}$ 
                                                                       \\
                                                                       & DSMIL\cite{dsmil}                & 331K   & 85.85$_{\pm5.11}$  & 81.17$_{\pm4.82}$  & 81.30$_{\pm4.63}$ &  & 97.96$_{\pm0.98}$ & 90.50$_{\pm2.97}$ & 92.06$_{\pm2.26}$ &  & 95.51$_{\pm0.55}$  & 80.61$_{\pm3.40}$ & 81.72$_{\pm2.79}$ 
                                                                       & &
                                                                       76.37$_{\pm2.01}$ &
                                                                       68.38$_{\pm3.01}$ &
                                                                       72.65$_{\pm2.68}$ 
                                                                       & &
                                                                       83.29$_{\pm4.08}$ &
                                                                       65.79$_{\pm5.54}$ &
                                                                       70.24$_{\pm5.98}$ 
                                                                       \\
                                                                       & RRT-MIL\cite{RRTMIL}              & 2.49M  & 94.60$_{\pm2.37}$ & 88.50$_{\pm3.13}$  & 88.53$_{\pm3.11}$ &  & 98.35$_{\pm0.38}$ & 90.99$_{\pm1.17}$ & 92.41$_{\pm0.75}$ &  & 95.58$_{\pm0.26}$  & 79.23$_{\pm3.26}$ & 80.25$_{\pm2.95}$ 
                                                                       & &
                                                                       78.77$_{\pm2.71}$ &
                                                                       69.22$_{\pm3.24}$ &
                                                                       73.58$_{\pm3.73}$ 
                                                                       & &
                                                                       84.79$_{\pm3.54}$ &
                                                                       68.02$_{\pm2.73}$ &
                                                                       70.98$_{\pm3.70}$
                                                                       \\
                                                                       & CoOp\cite{coop}                 & 341K   & 92.52$_{\pm2.53}$  & 84.97$_{\pm2.79}$  & 85.00$_{\pm2.76}$ &  & 98.03$_{\pm0.46}$ & 90.75$_{\pm0.66}$ & 92.23$_{\pm0.64}$ &  &  95.05$_{\pm0.92}$  & 78.58$_{\pm1.80}$ & 80.59$_{\pm1.78}$ 
                                                                       & &
                                                                       78.49$_{\pm2.92}$ &
                                                                       69.88$_{\pm2.79}$ &
                                                                       74.02$_{\pm2.74}$ 
                                                                       & &
                                                                       86.82$_{\pm1.62}$ &
                                                                       70.05$_{\pm1.88}$ &
                                                                       73.82$_{\pm1.18}$
                                                                       \\
                                                                       & CoCoOp\cite{cocoop}               & 383K   & 93.63$_{\pm2.26}$  & 86.73$_{\pm2.52}$  & 86.75$_{\pm2.50}$ &  & 97.92$_{\pm0.59}$ & 90.42$_{\pm1.05}$ & 91.98$_{\pm1.00}$ &  & 94.95$_{\pm0.59}$  & 78.92$_{\pm3.20}$ & 80.78$_{\pm2.70}$ 
                                                                       & &
                                                                       78.22$_{\pm2.90}$ &
                                                                       69.61$_{\pm2.40}$ &
                                                                       74.11$_{\pm2.52}$ 
                                                                       & &
                                                                       87.06$_{\pm1.02}$ &
                                                                       69.51$_{\pm2.03}$ &
                                                                       73.15$_{\pm1.01}$
                                                                       \\
                                                                       & Metaprompt\cite{metaprompt}           & 694K   & 94.98$_{\pm1.16}$  & 88.05$_{\pm0.94}$  & 88.08$_{\pm0.92}$ &  & 98.28$_{\pm0.39}$ & \uline{91.62$_{\pm0.98}$} & 92.51$_{\pm0.87}$ &  & 95.84$_{\pm0.55}$  & 80.59$_{\pm0.74}$ & 82.11$_{\pm0.83}$ 
                                                                       & &
                                                                       79.07$_{\pm3.22}$ &
                                                                       70.06$_{\pm3.64}$ &
                                                                       73.65$_{\pm4.25}$ 
                                                                       & &
                                                                       87.59$_{\pm0.48}$ &
                                                                       71.72$_{\pm0.44}$ &
                                                                       74.61$_{\pm0.35}$
                                                                       \\
                                                                       & TOP\cite{qu2024rise}                  & 2.76M  & 93.77$_{\pm 1.25}$       & 87.96$_{\pm 2.69}$      & 88.11$_{\pm 2.86}$   &  & 98.15$_{\pm 0.34}$       & 90.79$_{\pm 1.53}$      & 92.18$_{\pm 2.13}$     &  & 94.97$_{\pm 1.14}$       & 78.68$_{\pm 2.89}$      & 80.30$_{\pm 2.94}$  
                                                                       & &
                                                                       78.88$_{\pm3.37}$ &
                                                                       68.70$_{\pm3.94}$ &
                                                                       72.91$_{\pm4.72}$ 
                                                                       & &
                                                                       84.51$_{\pm2.11}$ &
                                                                       67.76$_{\pm2.85}$ &
                                                                       70.03$_{\pm3.33}$
                                                                       \\
                                                                       & ViLa-MIL\cite{shi2024vila}             & 2.32M  & 94.73$_{\pm1.76}$  & 87.58$_{\pm2.29}$  & 87.60$_{\pm2.30}$ &  & 98.00$_{\pm0.50}$ & 88.97$_{\pm1.93}$ & 90.54$_{\pm2.56}$ &  & 95.10$_{\pm1.03}$  & 79.90$_{\pm1.79}$ & 81.13$_{\pm1.79}$   
                                                                       & &
                                                                       78.31$_{\pm2.71}$ &
                                                                       69.80$_{\pm2.98}$ &
                                                                       74.34$_{\pm1.49}$ 
                                                                       & &
                                                                       85.22$_{\pm2.20}$ &
                                                                       69.04$_{\pm2.81}$ &
                                                                       71.68$_{\pm3.41}$ &
                                                                       \\
                                                   & \textbf{MSCPT(ours)} & 1.69M  & \uline{96.45$_{\pm0.94}$}  & \textbf{89.68$_{\pm2.71}$}  & \textbf{89.71$_{\pm2.70}$} &  & \textbf{98.66$_{\pm0.29}$} & \textbf{92.50$_{\pm1.18}$} & \textbf{93.45$_{\pm1.16}$} &  & \uline{96.14$_{\pm0.79}$}  & \textbf{83.05$_{\pm2.04}$} & \textbf{83.58$_{\pm1.94}$}  
                                                   & &
                                                   \textbf{80.81$_{\pm3.10}$} &
                                                   \textbf{71.71$_{\pm2.79}$} &
                                                   \textbf{75.86$_{\pm2.41}$} 
                                                   & &
                                                   \textbf{89.39$_{\pm1.71}$} &
                                                   \textbf{73.58$_{\pm2.84}$} &
                                                   \textbf{76.66$_{\pm2.61}$} \\

\end{tblr}
\label{table1}
}
\vspace{-1.5em}
\end{table*}

\section{Experimental Results}
\subsection{Experimental Settings}
\subsubsection{Datasets}
To comprehensively evaluate MSCPT's performance, we used five real-world datasets across three downstream tasks, collected from multiple centers. Additionally, to assess the generalizability of MSCPT across different VLMs, we conducted extensive experiments using three selected VLMs.

\textbf{TCGA-NSCLS}, \textbf{TCGA-RCC}, and \textbf{UBC-OCEAN} datasets are used to examine the cancer subtyping ability of MSCPT. \textbf{TCGA-NSCLC} contains 1041 non-small cell lung cancer WSIs, comprising 530 lung adenocarcinoma (LUAD) and 511 lung squamous cell carcinoma (LUSC) WSIs.
\textbf{TCGA-RCC} includes 873 renal cell carcinoma WSIs: 121 chromophobe renal cell carcinoma (CHRCC), 455 clear-cell renal cell carcinoma (CCRCC), and 297 papillary renal cell carcinoma (PRCC).
\textbf{UBC-OCEAN}~\cite{UBC-OCEAN} comprises 538 ovarian WSIs, with 99 clear cell carcinoma (CC), 124 endometrioid carcinoma (EC), 221 high-grade serous carcinoma (HGSC), 47 low-grade serous carcinoma (LGSC), and 46 mucinous carcinoma (MC) WSIs.

The \textbf{TCGA-BRCA} dataset is used for predicting the risk of cancer recurrence and comprises 1,099 WSIs. Typically, the recurrence risk of breast cancer is predicted by gene expression data combined with OncotypeDx (ODX). Although the original dataset does not provide labels related to recurrence risk, Howard~\textit{et al.}~\cite{howard2023integration} calculated the corresponding ODX scores based on the mRNA expression data provided by TCGA and classified the patients into high recurrence risk (316 WSIs) and low recurrence risk (783 WSIs).

The \textbf{PANDA} dataset includes 10,616 WSIs used for prostate cancer grading. Each slide is labeled as normal or assigned an International Society of Urological Pathology (ISUP) grade, resulting in six classes. Considering the difficulty of the few-shot weakly supervised WSI classification task, we introduced coarse-grained labels based on European Association of Urology (EAU) prostate cancer guidelines~\cite{salonia2023eau}, mapping ISUP grades to risk categories: normal or grade 1 as low risk (5,558 WSIs), grade 2 and grade 3 as medium risk (2,585 WSIs), and grades 4–5 as high risk (2,473 WSIs).

\subsubsection{Evaluation Metrics}
For all datasets, we leverage Accuracy (ACC), Area Under Curve (AUC), and macro F1-score (F1) to evaluate model performance. To reduce the impact of data split on model evaluation, we employ five fixed seeds to perform five rounds of dataset splitting, model training, and testing. We report the mean and standard deviation of the metrics. 

\subsubsection{Model Zoo}
Thirteen influential approaches were employed for comparison, including traditional MIL-based methods: Mean pooling, Max pooling, ABMIL~\cite{abmil}, CLAM~\cite{clam}, TransMIL~\cite{shao2021transmil}, DSMIL~\cite{dsmil} and RRT-MIL~\cite{RRTMIL}; prompt tuning methods for natural images: CoOp~\cite{coop}, CoCoOp~\cite{cocoop} and Metaprompt~\cite{metaprompt}; prompt tuning methods for WSIs: TOP~\cite{qu2024rise} and ViLa-MIL~\cite{shi2024vila}. Adapting to WSI-level tasks, we integrated an attention-based instance aggregation module~\cite{abmil} into the prompt tuning methods designed for natural images. Additionally, the prompts for CoOp, CoCoOp, and MetaPrompt were set according to the format provided in their official code: \textit{``\textbf{\textbf{class name}}"} or \textit{``a photo of {\textbf{class name}}"}.

\subsubsection{Implementation Details}
Following CLAM~\cite{clam}, the original WSIs were initially processed using the Otsu thresholding algorithm to remove the background parts. Subsequently, the WSIs were segmented into multiple non-overlapping patches of $256\times256$ pixels at $5\times$ and $20\times$ magnification levels. We evaluate MSCPT on CLIP~\cite{clip}, PLIP~\cite{plip}, and CONCH~\cite{conch}, with the first two using ViT-B/32~\cite{vit} and CONCH using ViT-B/16~\cite{vit} as their visual towers. Apart from MSCPT, Metaprompt, and DSMIL, which used both $5\times$ and $20\times$ magnification patches, the remaining methods solely relied on $20\times$ magnification patches as inputs. To ensure fair comparisons across different shot settings, 80\% of each dataset was used for testing, while $K$ samples per class ($K=16,8,4,2$) were drawn from the remaining 20\% for training.

All methods used the Adam optimizer (learning rate: $1e-4$, weight decay: $1e-5$) with a batch size of 1. Training was fixed at 100 epochs for CLIP and 50 for PLIP and CONCH, with early stopping applied. We chose GPT-4~\cite{gpt4} to generate pathological visual descriptions, providing 10 low-level and 30 high-level visual descriptions for each category of WSIs (\textit{i.e.}, $C^{l}=10$ and $C^{h}=30$). For MSCPT and Metaprompt, we selected 30 patches for each category at $5\times$ magnification for visual prompt tuning. The lengths of the global prompts $p_{glob}$ in both image and text encoders were uniformly set to 2. All experiments were conducted on a workstation with eight NVIDIA A800 GPUs.

\begin{table*}[t]
\centering
\caption{Results on five datasets, three downstream tasks under the 2-shot, 4-shot and 8-shot setting using CONCH. The highest performance is in bold, the second-best underlined, with mean and standard deviation reported across five random seeds.}
\resizebox{\textwidth}{!}{
\begin{tblr}{
  colsep=1.5pt,
  rowsep = 0.05cm,
  % column{2} = {l},
  column{3} = {c},
  column{4} = {c},
  column{5} = {c},
  column{6} = {c},
  column{7} = {c},
  column{8} = {c},
  column{9} = {c},
  column{10} = {c},
  column{11} = {c},
  column{12} = {c},
  column{13} = {c},
  column{14} = {c},
  column{15} = {c},
  column{16} = {c},
  column{17} = {c},
  column{18} = {c},
  column{19} = {c},
  column{20} = {c},
  column{21} = {c},
  column{22} = {c},
  column{23} = {c},
  row{16} = {bg=lightgray},
  row{30} = {bg=lightgray},
  row{44} = {bg=lightgray},
  cell{1}{1} = {c=2,r=2}{c},
  % cell{1}{3} = {r=2}{},
  cell{1}{3} = {c=3}{},
  cell{1}{7} = {c=3}{},
  cell{1}{11} = {c=3}{},
  cell{1}{15} = {c=3}{},
  cell{1}{19} = {c=3}{},
  cell{3}{1} = {r=14}{c, bg=lightgray},
  cell{17}{1} = {r=14}{c, bg=lightgray},
  cell{31}{1} = {r=14}{c, bg=lightgray},
  hline{1,17,31,45} = {-}{0.08em},
  hline{2} = {3-5,7-9,11-13,15-17,19-21}{},
  hline{3} = {-}{},
}
                                                                \textbf{Methods}      & & {\textbf{TCGA-NSCLC} \\ (Subtyping, 2 classes)}  & &  ～      & & {\textbf{TCGA-RCC} \\ (Subtyping, 3 classes)}  &            &     ～ &  & {\textbf{UBC-OCEAN} \\ (Subtyping, 5 classes)}     & &   ～     & & {\textbf{TCGA-BRCA} \\ (Recurrence Prediction, 2 classes)}     & &    ～    & & {\textbf{PANDA} \\ (Grading, 3 classes)}         \\
                                                                     &  & AUC         & F1          & ACC        &  & AUC         & F1         & ACC         &  & AUC         & F1         & ACC  &  & AUC        & F1         & ACC &  & AUC        & F1         & ACC      \\
\begin{sideways}\makecell{\large \textbf{2-shot} \\ 2 Training Samples per Class}\end{sideways}              
                                                                       & \raggedright Max-pooling         & 80.85$_{\pm5.19}$  & 64.68$_{\pm6.49}$  & 67.31$_{\pm4.22}$ &  & 86.62$_{\pm5.10}$ & 69.08$_{\pm8.05}$ & 73.21$_{\pm7.63}$ &  & 84.82$_{\pm2.76}$  & 58.70$_{\pm5.87}$ & 62.65$_{\pm5.75}$ 
                                                                       & &
                                                                       67.92$_{\pm6.34}$ &
                                                                       61.66$_{\pm7.31}$ &
                                                                       67.60$_{\pm5.07}$ 
                                                                       & &
                                                                       70.31$_{\pm3.67}$ &
                                                                       51.20$_{\pm4.20}$ &
                                                                       56.44$_{\pm3.67}$ &
                                                                       \\
                                                                       & \raggedright Mean-pooling       & 78.29$_{\pm5.24}$  & 70.35$_{\pm4.44}$  & 70.79$_{\pm4.32}$ &  & 93.08$_{\pm2.79}$ & 77.45$_{\pm6.52}$ & 80.63$_{\pm5.33}$ &  & 85.69$_{\pm2.97}$  & 57.82$_{\pm5.36}$ & 59.66$_{\pm8.76}$ & &66.34$_{\pm8.16}$  & 58.96$_{\pm8.70}$ & 62.57$_{\pm7.07}$ 
                                                                       & &
                                                                       71.59$_{\pm5.06}$ &
                                                                       51.57$_{\pm4.66}$ &
                                                                       56.66$_{\pm5.60}$ &
                                                                       \\
                                                                       & \raggedright ABMIL~\cite{abmil}   & 81.08$_{\pm7.36}$  & 71.12$_{\pm6.84}$  & 71.68$_{\pm6.55}$ &  & 89.84$_{\pm5.32}$ & 74.71$_{\pm5.71}$ & 79.63$_{\pm5.10}$ &  & 87.06$_{\pm3.94}$  & 61.49$_{\pm7.69}$ & \uline{64.61$_{\pm7.85}$}
                                                                       & &
                                                                \uline{70.38$_{\pm3.82}$} &
                                                                       62.27$_{\pm4.48}$ &
                                                                68.92$_{\pm3.89}$ 
                                                                       & &
                                                                       71.48$_{\pm6.09}$ &
                                                                       52.36$_{\pm4.95}$ &
                                                                       57.99$_{\pm5.11}$ &
                                                                       \\
                                                                       & \raggedright CLAM-SB~\cite{clam}  & 83.73$_{\pm5.36}$  & 72.67$_{\pm3.61}$  & 73.25$_{\pm3.26}$ &  & 93.05$_{\pm3.01}$ & 79.99$_{\pm6.52}$ & 83.99$_{\pm5.75}$ &  & \uline{88.11$_{\pm4.40}$} & \uline{62.13$_{\pm4.53}$} & 62.84$_{\pm4.14}$  
                                                                       & &
                                                                       67.45$_{\pm8.24}$ &
                                                                       60.25$_{\pm7.22}$ &
                                                                       64.73$_{\pm8.64}$
                                                                       & &
                                                                       71.16$_{\pm6.91}$ &
                                                                       50.69$_{\pm9.72}$ &
                                                                       54.50$_{\pm12.12}$ 
                                                                       \\
                                                                       & \raggedright CLAM-MB~\cite{clam}  & 84.35$_{\pm7.86}$  & 73.44$_{\pm6.68}$  & 74.16$_{\pm6.19}$ &  & 93.14$_{\pm2.33}$ & 79.91$_{\pm7.63}$ & 82.35$_{\pm6.36}$ & & 87.21$_{\pm4.01}$  & 59.93$_{\pm6.25}$ & 61.18$_{\pm8.45}$ 
                                                                       & &
                                                                       68.28$_{\pm6.75}$ &
                                                                       62.05$_{\pm5.60}$ &
                                                                       66.32$_{\pm6.60}$ 
                                                                       & &
                                                                       72.66$_{\pm4.86}$ &
                                                                       52.08$_{\pm4.92}$ &
                                                                       56.73$_{\pm6.12}$
                                                                       \\
                                                                       & \raggedright TransMIL~\cite{shao2021transmil} & 78.97$_{\pm5.95}$  & 71.34$_{\pm5.45}$ & 71.63$_{\pm5.36}$ &  & 93.70$_{\pm2.84}$ & 76.99$_{\pm1.52}$ & 79.74$_{\pm5.88}$ &  & 86.64$_{\pm3.58}$  & 59.42$_{\pm5.64}$ & 61.23$_{\pm7.68}$ 
                                                                       & &
                                                                       64.51$_{\pm4.71}$ &
                                                                       60.88$_{\pm6.05}$ &
                                                                       67.30$_{\pm5.04}$ 
                                                                       & &
                                                                       70.53$_{\pm6.26}$ &
                                                                       49.80$_{\pm5.97}$ &
                                                                       54.90$_{\pm6.86}$
                                                                       \\
                                                                       & \raggedright DSMIL~\cite{dsmil}   & 73.83$_{\pm9.24}$  & 69.19$_{\pm8.07}$  & 69.52$_{\pm7.77}$ &  & 80.94$_{\pm5.33}$ & 62.41$_{\pm6.18}$ & 69.43$_{\pm7.06}$ &  & 80.61$_{\pm5.31}$ & 52.24$_{\pm8.83}$ & 55.00$_{\pm8.16}$
                                                                       & &
                                                                       63.61$_{\pm7.31}$ &
                                                                       61.11$_{\pm6.67}$ &
                                                                       65.22$_{\pm6.75}$ 
                                                                       & &
                                                                       62.30$_{\pm7.21}$ &
                                                                       45.78$_{\pm8.04}$ &
                                                                       49.11$_{\pm10.33}$ &
                                                                       \\
                                                                       & \raggedright RRT-MIL~\cite{RRTMIL} & 80.74$_{\pm7.70}$  & 74.37$_{\pm6.25}$  & 74.40$_{\pm6.22}$ &  & 92.03$_{\pm2.55}$ & 77.33$_{\pm6.77}$ & 79.60$_{\pm6.26}$ &  & 85.75$_{\pm3.78}$  & 57.92$_{\pm3.49}$ & 58.04$_{\pm5.82}$
                                                                       & &
                                                                       68.57$_{\pm5.50}$ &
                                                                       62.48$_{\pm6.47}$ &
                                                                \uline{69.08$_{\pm2.24}$} 
                                                                       & &
                                                                       70.09$_{\pm8.40}$ &
                                                                       48.42$_{\pm6.94}$ &
                                                                       54.93$_{\pm6.57}$ &
                                                                       \\
                                                                       & \raggedright CoOp~\cite{coop}    & 76.93$_{\pm6.08}$  & 69.93$_{\pm5.21}$  & 70.12$_{\pm5.14}$  &  & 91.62$_{\pm3.38}$ & 75.58$_{\pm6.00}$ & 78.14$_{\pm5.29}$ &  & 82.78$_{\pm5.85}$  & 53.18$_{\pm8.62}$ & 55.00$_{\pm10.26}$  
                                                                       & &
                                                                       67.71$_{\pm5.68}$ &
                                                                       59.62$_{\pm4.55}$ &
                                                                       64.25$_{\pm4.63}$ 
                                                                       & &
                                                                       72.08$_{\pm6.49}$ &
                                                                       49.33$_{\pm6.59}$ &
                                                                       56.03$_{\pm5.80}$ &
                                                                       \\
                                                                       & \raggedright CoCoOp~\cite{cocoop} & 79.63$_{\pm5.73}$  & 71.87$_{\pm5.44}$  & 72.19$_{\pm5.25}$ &  & 93.06$_{\pm2.78}$ & 77.79$_{\pm6.88}$ & 80.77$_{\pm6.02}$ &  & 85.62$_{\pm1.97}$ & 59.68$_{\pm3.12}$ & 63.28$_{\pm4.47}$ 
                                                                       & &
                                                                       66.17$_{\pm7.28}$ &
                                                                \uline{62.89$_{\pm6.71}$} &
                                                                       66.33$_{\pm7.46}$ 
                                                                       & &
                                                                       72.30$_{\pm4.68}$ &
                                                                       49.31$_{\pm2.67}$ &
                                                                       53.66$_{\pm6.24}$ 
                                                                       \\
                                                                       & \raggedright Metaprompt~\cite{metaprompt} & \uline{87.11$_{\pm4.37}$}  & 
                                                                \uline{78.30$_{\pm5.21}$}  & 
                                                                \uline{78.51$_{\pm4.99}$}             &  & 
                                                                \uline{94.26$_{\pm2.58}$} & \uline{82.45$_{\pm4.68}$} & \uline{84.58$_{\pm4.12}$} 
                                                                       & & 
                                                                       85.86$_{\pm2.89}$ & 58.80$_{\pm5.86}$ & 60.64$_{\pm7.29}$   
                                                                       & &
                                                                       67.61$_{\pm4.87}$ &
                                                                       61.78$_{\pm4.86}$ &
                                                                       67.35$_{\pm5.57}$ 
                                                                       & &
                                                                    \uline{75.09$_{\pm4.60}$} &
                                                                    \uline{53.43$_{\pm3.45}$} &
                                                                    \uline{58.44$_{\pm3.99}$}
                                                                       \\
                                                                       & \raggedright TOP~\cite{qu2024rise} & 81.46$_{\pm 5.32}$      & 75.62$_{\pm 4.88}$    & 76.01$_{\pm 4.92}$    &  & 
                                                                       92.14$_{\pm 2.77}$      & 76.50$_{\pm 5.38}$    & 77.14$_{\pm 5.20}$  
                                                                       &  & 
                                                                       84.24$_{\pm 4.00}$      & 56.38$_{\pm 3.82}$    & 57.41$_{\pm 3.45}$ 
                                                                       & &
                                                                       65.91$_{\pm 7.05}$ &
                                                                       60.24$_{\pm 4.06}$ &
                                                                       67.02$_{\pm 4.46}$ 
                                                                       & &
                                                                       70.33$_{\pm7.14}$ &
                                                                       51.87$_{\pm6.46}$ &
                                                                       56.58$_{\pm5.11}$
                                                                       \\
                                                                       & \raggedright ViLa-MIL~\cite{shi2024vila}  & 84.31$_{\pm6.52}$  & 76.23$_{\pm4.74}$  & 76.39$_{\pm4.78}$ &  & 92.20$_{\pm3.90}$ & 71.47$_{\pm6.12}$ & 73.55$_{\pm5.94}$ &  &  85.95$_{\pm3.74}$  & 58.55$_{\pm3.12}$ & 59.71$_{\pm5.02}$  
                                                                       & &
                                                                       62.35$_{\pm9.85}$ &
                                                                       57.37$_{\pm8.52}$ &
                                                                       66.26$_{\pm7.03}$ 
                                                                       & &
                                                                       71.92$_{\pm6.81}$ &
                                                                       52.52$_{\pm5.51}$ &
                                                                       58.11$_{\pm4.24}$ &
                                                                       \\
                                                                      & \raggedright \textbf{MSCPT(ours)} & \textbf{93.78$_{\pm1.21}$}  & \textbf{84.38$_{\pm3.68}$}  & \textbf{84.47$_{\pm3.57}$}  &  & \textbf{95.12$_{\pm1.91}$} & \textbf{83.09$_{\pm5.16}$} & \textbf{86.47$_{\pm3.91}$} &  & \textbf{89.68$_{\pm3.83}$}  & \textbf{64.16$_{\pm5.70}$} &\textbf{68.33$_{\pm5.55}$}   
                                                                      & &
                                                                \textbf{72.06$_{\pm4.28}$} &
                                                                \textbf{64.58$_{\pm3.40}$} &
                                                                \textbf{69.46$_{\pm3.40}$} 
                                                                & &
                                                                \textbf{76.19$_{\pm6.18}$} &
                                                                \textbf{56.82$_{\pm4.13}$} &
                                                                \textbf{61.81$_{\pm4.79}$} &
                                                                      \\
\begin{sideways}\makecell{\large \textbf{4-shot} \\ 4 Training Samples per Class}\end{sideways}
                                                                       & \raggedright Max-pooling       & 84.75$_{\pm4.38}$   & 76.52$_{\pm5.40}$  & 76.66$_{\pm5.38}$ &  & 92.80$_{\pm2.37}$ & 77.72$_{\pm5.20}$ & 80.89$_{\pm4.58}$ &  & 90.18$_{\pm1.52}$  & 68.34$_{\pm6.33}$ & 69.85$_{\pm4.20}$  
                                                                       & &
                                                                       72.55$_{\pm6.54}$ &
                                                                       64.69$_{\pm4.66}$ &
                                                                       68.95$_{\pm2.82}$
                                                                       & &
                                                                       75.09$_{\pm2.06}$ &
                                                                       56.71$_{\pm2.99}$ &
                                                                       61.33$_{\pm3.18}$ &
                                                                       \\
                                                                       & \raggedright Mean-pooling   & 79.66$_{\pm6.71}$  & 73.00$_{\pm4.89}$  & 73.05$_{\pm4.91}$ &  & 93.57$_{\pm2.07}$ & 78.94$_{\pm4.86}$ & 81.81$_{\pm4.02}$ &  & 88.32$_{\pm0.96}$  & 63.18$_{\pm2.70}$ & 66.91$_{\pm3.56}$  
                                                                       & &
                                                                       71.78$_{\pm6.43}$ &
                                                                       63.29$_{\pm4.39}$ &
                                                                       67.97$_{\pm3.60}$ 
                                                                       & &
                                                                       76.03$_{\pm4.55}$ &
                                                                       57.11$_{\pm3.99}$ &
                                                                       62.80$_{\pm3.79}$ &
                                                                       \\
                                                                       & \raggedright ABMIL~\cite{abmil}  & 84.54$_{\pm7.61}$  & 75.71$_{\pm7.44}$  & 75.89$_{\pm7.30}$ &  & 93.01$_{\pm4.97}$ & 81.29$_{\pm4.20}$ & 84.56$_{\pm3.22}$ &  & 91.72$_{\pm2.13}$  & 69.60$_{\pm7.34}$ & 72.79$_{\pm7.44}$ 
                                                                       & &
                                                                \uline{74.59$_{\pm6.82}$} &
                                                                       64.16$_{\pm6.97}$ &
                                                                       68.87$_{\pm6.51}$ 
                                                                       & &
                                                                       79.17$_{\pm4.97}$ &
                                                                       59.48$_{\pm3.45}$ &
                                                                       64.98$_{\pm4.87}$ &
                                                                       \\
                                                                       & \raggedright CLAM-SB~\cite{clam}  & 87.01$_{\pm7.93}$  & 78.19$_{\pm8.65}$  & 78.63$_{\pm8.13}$ &  & 96.12$_{\pm1.39}$ & 84.49$_{\pm3.30}$ & 86.38$_{\pm3.31}$ &  & 90.55$_{\pm1.29}$  & 67.30$_{\pm5.63}$ & 69.80$_{\pm5.35}$  
                                                                       & &
                                                                       72.73$_{\pm9.88}$ &
                                                                       64.42$_{\pm5.97}$ &
                                                                       69.49$_{\pm5.55}$ 
                                                                       & &
                                                                       79.31$_{\pm4.32}$ &
                                                                       \uline{61.87$_{\pm3.27}$} &
                                                                       \uline{66.97$_{\pm3.87}$} &
                                                                       \\
                                                                       & \raggedright CLAM-MB~\cite{clam}  & 86.19$_{\pm10.61}$  & 78.71$_{\pm10.68}$  & 78.92$_{\pm10.44}$ &  &
                                                                \uline{96.24$_{\pm2.11}$} & 
                                                                \uline{84.69$_{\pm4.83}$} & 
                                                                \uline{86.93$_{\pm4.27}$} &  & 
                                                                \uline{92.04$_{\pm1.41}$}  & \uline{70.72$_{\pm4.94}$} & 72.55$_{\pm4.95}$  
                                                                       & &
                                                                       73.20$_{\pm7.79}$ &
                                                                       65.08$_{\pm5.97}$ &
                                                                       71.15$_{\pm3.26}$ 
                                                                       & &
                                                                       79.63$_{\pm4.13}$ &
                                                                       60.80$_{\pm3.68}$ &
                                                                       65.32$_{\pm4.55}$ &
                                                                       \\
                                                                       & \raggedright TransMIL~\cite{shao2021transmil}         & 81.71$_{\pm10.75}$  & 74.87$_{\pm9.06}$  & 74.95$_{\pm8.99}$ &  & 95.30$_{\pm1.69}$ & 81.59$_{\pm3.75}$ & 83.93$_{\pm3.04}$ &  & 90.12$_{\pm1.71}$  & 66.11$_{\pm6.50}$ & 68.58$_{\pm6.63}$
                                                                       & &
                                                                       71.88$_{\pm6.41}$ &
                                                                       65.17$_{\pm4.17}$ &
                                                                       71.95$_{\pm2.79}$ 
                                                                       & &
                                                                       77.10$_{\pm4.32}$ &
                                                                       58.06$_{\pm4.49}$ &
                                                                       63.31$_{\pm4.82}$ &
                                                                       \\
                                                                       & \raggedright DSMIL~\cite{dsmil}   & 76.42$_{\pm12.49}$  & 72.49$_{\pm9.98}$  & 72.62$_{\pm10.00}$ &  & 92.00$_{\pm5.12}$ & 77.37$_{\pm9.20}$ & 82.52$_{\pm6.56}$ &  & 85.84$_{\pm7.61}$  & 60.82$_{\pm13.07}$ & 62.99$_{\pm14.10}$  
                                                                       & &
                                                                       67.63$_{\pm6.40}$ &
                                                                       63.63$_{\pm6.65}$ &
                                                                       68.44$_{\pm5.05}$ 
                                                                       & &
                                                                       74.37$_{\pm5.24}$ &
                                                                       55.33$_{\pm3.78}$ &
                                                                       61.85$_{\pm4.99}$ 
                                                                       \\
                                                                       & \raggedright RRT-MIL~\cite{RRTMIL} & 83.05$_{\pm10.92}$ & 77.08$_{\pm9.70}$  & 77.09$_{\pm9.70}$ &  & 95.03$_{\pm1.76}$ & 81.67$_{\pm6.59}$ & 83.87$_{\pm6.41}$ &  & 89.88$_{\pm1.23}$  & 64.75$_{\pm6.01}$ & 66.42$_{\pm6.24}$  
                                                                       & &
                                                                       74.53$_{\pm4.77}$ &
                                                                       65.63$_{\pm3.80}$ &
                                                                       71.65$_{\pm3.48}$ 
                                                                       & &
                                                                       77.86$_{\pm4.05}$ &
                                                                       59.19$_{\pm3.66}$ &
                                                                       64.31$_{\pm5.19}$ 
                                                                       \\
                                                                       & \raggedright CoOp~\cite{coop}   & 80.83$_{\pm9.18}$  & 73.87$_{\pm8.19}$  & 74.01$_{\pm8.06}$ &  & 93.96$_{\pm2.41}$ & 81.07$_{\pm3.80}$ & 83.21$_{\pm3.95}$ &  &  89.63$_{\pm1.31}$  & 65.79$_{\pm5.44}$ & 69.12$_{\pm5.27}$  
                                                                       & &
                                                                       71.15$_{\pm8.50}$ &
                                                                       64.66$_{\pm5.37}$ &
                                                                       69.64$_{\pm3.60}$ 
                                                                       & &
                                                                       79.17$_{\pm2.91}$ &
                                                                       56.21$_{\pm6.55}$ &
                                                                       62.28$_{\pm6.25}$
                                                                       \\
                                                                       & \raggedright CoCoOp~\cite{cocoop} & 80.46$_{\pm9.40}$  & 73.85$_{\pm8.11}$  & 73.92$_{\pm8.09}$ &  & 93.96$_{\pm2.95}$ & 81.07$_{\pm3.80}$ & 83.21$_{\pm3.95}$ &  & 88.44$_{\pm4.23}$  & 61.04$_{\pm9.45}$ & 64.90$_{\pm11.03}$   
                                                                       & &
                                                                       71.11$_{\pm7.96}$ &
                                                                       65.02$_{\pm5.90}$ &
                                                                       68.04$_{\pm6.43}$ 
                                                                       & &
                                                                       78.70$_{\pm4.10}$ &
                                                                       57.06$_{\pm3.31}$ &
                                                                       63.13$_{\pm4.85}$
                                                                       \\
                                                                       & \raggedright Metaprompt~\cite{metaprompt} & 
                                                                \uline{87.59$_{\pm5.98}$}  & \uline{79.69$_{\pm5.82}$}  & \uline{79.78$_{\pm5.75}$} &  & 95.98$_{\pm1.55}$ & 84.20$_{\pm3.04}$ & 86.48$_{\pm3.06}$ &  & 91.42$_{\pm2.56}$  & 67.95$_{\pm5.18}$ & \uline{72.99$_{\pm2.97}$} 
                                                                       & &
                                                                       72.72$_{\pm5.47}$ &
                                                                       
                                                                \uline{66.50$_{\pm3.74}$} &
                                                                \uline{72.31$_{\pm3.45}$} 
                                                                       & &
                                                                    \uline{81.48$_{\pm2.27}$} &
                                                                    59.96$_{\pm7.24}$ &
                                                                    65.97$_{\pm6.02}$ &
                                                                       \\
                                                                       & \raggedright TOP~\cite{qu2024rise} & 84.49$_{\pm 7.66}$      & 77.50$_{\pm 7.51}$    & 77.53$_{\pm 7.32}$    &  & 
                                                                       95.01$_{\pm 1.44}$      & 82.46$_{\pm 3.87}$    & 84.70$_{\pm 3.61}$  
                                                                       &  & 
                                                                       90.45$_{\pm 1.11}$      & 67.89$_{\pm 4.05}$    & 69.04$_{\pm 3.72}$ 
                                                                       & &
                                                                       72.65$_{\pm 6.10}$ &
                                                                       65.66$_{\pm 4.97}$ &
                                                                       68.18$_{\pm 5.78}$
                                                                       & &
                                                                       78.43$_{\pm4.72}$ &
                                                                       58.44$_{\pm2.19}$ &
                                                                       64.39$_{\pm2.76}$ &
                                                                       \\
                                                                       & \raggedright ViLa-MIL~\cite{shi2024vila}  & 85.67$_{\pm8.20}$  & 78.15$_{\pm8.19}$  & 78.20$_{\pm8.13}$ &  & 94.65$_{\pm1.72}$ & 79.24$_{\pm4.67}$ & 82.09$_{\pm4.74}$ &  & 89.72$_{\pm0.90}$  & 63.88$_{\pm3.86}$ & 66.23$_{\pm3.24}$   
                                                                       & &
                                                                       70.21$_{\pm8.80}$ &
                                                                       62.72$_{\pm5.51}$ &
                                                                       68.44$_{\pm6.17}$ 
                                                                       & &
                                                                       76.01$_{\pm5.10}$ &
                                                                       57.83$_{\pm3.56}$ &
                                                                       62.85$_{\pm3.60}$
                                                                       \\
                                                   & \raggedright \textbf{MSCPT(ours)} & \textbf{93.60$_{\pm2.76}$}  & \textbf{86.13$_{\pm3.58}$}  & \textbf{86.15$_{\pm3.54}$} &  & \textbf{97.03$_{\pm2.05}$} & \textbf{87.26$_{\pm5.03}$} & \textbf{89.48$_{\pm4.02}$} &  & \textbf{93.19$_{\pm1.97}$}  & \textbf{73.53$_{\pm5.26}$} & \textbf{76.81$_{\pm5.10}$}     
                                                   & &
                                                   \textbf{76.20$_{\pm4.17}$} &
                                                   \textbf{68.25$_{\pm4.82}$} &
                                                   \textbf{73.60$_{\pm3.00}$} 
                                                   & &
                                                   \textbf{82.33$_{\pm3.51}$} &
                                                   \textbf{64.13$_{\pm1.51}$} &
                                                   \textbf{68.41$_{\pm2.46}$} 
                                                   \\
\begin{sideways}\makecell{\large \textbf{8-shot} \\ 8 Training Samples per Class}\end{sideways}
                                                                       & \raggedright Max-pooling    & 91.54$_{\pm1.81}$   & 83.41$_{\pm3.23}$  & 83.46$_{\pm3.16}$ &  & 97.39$_{\pm0.54}$ & 88.47$_{\pm1.53}$ & 90.17$_{\pm1.10}$ &  & 91.72$_{\pm1.15}$  & 71.43$_{\pm1.55}$ & 73.84$_{\pm2.08}$  
                                                                       & &
                                                                       74.62$_{\pm1.91}$ &
                                                                       66.56$_{\pm1.69}$ &
                                                                       69.70$_{\pm2.10}$ 
                                                                       & &
                                                                       77.49$_{\pm4.34}$ &
                                                                       59.38$_{\pm4.25}$ &
                                                                       64.19$_{\pm4.15}$ &
                                                                       \\
                                                                       & \raggedright Mean-pooling     & 85.37$_{\pm2.94}$  & 77.42$_{\pm3.52}$  & 77.52$_{\pm3.41}$ &  & 96.61$_{\pm0.38}$ & 86.26$_{\pm1.27}$ & 88.34$_{\pm1.18}$ &  & 91.83$_{\pm1.15}$  & 70.48$_{\pm2.87}$ & 72.79$_{\pm3.43}$   
                                                                       & &
                                                                       72.47$_{\pm2.84}$ &
                                                                       64.63$_{\pm2.16}$ &
                                                                       68.65$_{\pm1.90}$
                                                                       & &
                                                                       78.07$_{\pm4.34}$ &
                                                                       60.21$_{\pm5.46}$ &
                                                                       64.49$_{\pm3.92}$ &
                                                                       \\
                                                                       & \raggedright ABMIL~\cite{abmil}  & 88.60$_{\pm8.50}$  & 80.73$_{\pm8.51}$  & 80.82$_{\pm8.37}$ &  & 97.44$_{\pm1.96}$ & 89.08$_{\pm5.62}$ & 90.46$_{\pm5.00}$ &  & 94.43$_{\pm0.96}$  & 
                                                                76.33$_{\pm2.16}$ & \uline{78.63$_{\pm1.93}$}  
                                                                       & &
                                                                       75.37$_{\pm4.03}$ &
                                                                       66.16$_{\pm4.17}$ &
                                                                       72.49$_{\pm6.70}$ 
                                                                       & &
                                                                       83.37$_{\pm4.34}$ &
                                                                       65.11$_{\pm5.66}$ &
                                                                       68.79$_{\pm5.49}$ 
                                                                       \\
                                                                       & \raggedright CLAM-SB~\cite{clam}  & 
                                                                \uline{92.62$_{\pm5.11}$}  & \uline{85.92$_{\pm5.76}$}  & \uline{85.94$_{\pm5.73}$} &  & 97.56$_{\pm0.28}$ & \uline{90.87$_{\pm0.88}$} & 91.12$_{\pm0.59}$ &  & \uline{94.44$_{\pm0.74}$}  & \uline{77.12$_{\pm1.51}$} & 78.22$_{\pm1.96}$   
                                                                       & &
                                                                       
                                                                \uline{77.02$_{\pm4.66}$} &
                                                                       67.96$_{\pm4.58}$ &
                                                                       73.02$_{\pm3.11}$ 
                                                                       & &
                                                                    \uline{83.93$_{\pm2.82}$} &
                                                                    \uline{65.67$_{\pm4.42}$} &
                                                                    \uline{69.67$_{\pm3.95}$} 
                                                                       \\
                                                                       & \raggedright CLAM-MB~\cite{clam}    & 91.99$_{\pm5.17}$  & 84.10$_{\pm6.99}$  & 84.18$_{\pm6.85}$ &  & 97.39$_{\pm0.43}$ & 89.64$_{\pm1.60}$ & 90.55$_{\pm1.11}$ &  & 
                                                                94.36$_{\pm0.85}$  & 76.19$_{\pm1.65}$ & 77.65$_{\pm2.26}$  
                                                                       & &
                                                                       76.94$_{\pm4.36}$ &
                                                                       67.94$_{\pm4.93}$ &
                                                                       
                                                                \uline{73.56$_{\pm3.08}$} 
                                                                       & &
                                                                       82.63$_{\pm3.73}$ &
                                                                       63.64$_{\pm5.41}$ &
                                                                       67.49$_{\pm4.92}$
                                                                       \\
                                                                       & \raggedright TransMIL~\cite{shao2021transmil}         & 88.71$_{\pm2.96}$  & 80.81$_{\pm3.94}$  & 80.91$_{\pm3.82}$ &  & 97.97$_{\pm0.58}$ & 89.78$_{\pm1.09}$ & 91.32$_{\pm0.93}$ &  & 93.53$_{\pm0.62}$  & 72.97$_{\pm2.27}$ & 75.05$_{\pm3.31}$  
                                                                       & &
                                                                       75.39$_{\pm2.61}$ &
                                                                       67.52$_{\pm2.16}$ &
                                                                       72.17$_{\pm2.39}$ 
                                                                       & &
                                                                       80.68$_{\pm4.10}$ &
                                                                       62.45$_{\pm5.10}$ &
                                                                       66.92$_{\pm4.77}$
                                                                       \\
                                                                       & \raggedright DSMIL~\cite{dsmil} & 79.79$_{\pm10.05}$  & 75.01$_{\pm9.41}$  & 75.05$_{\pm9.39}$ &  & 95.18$_{\pm5.52}$ & 84.89$_{\pm9.80}$ & 87.97$_{\pm7.62}$ &  & 92.72$_{\pm1.03}$  & 73.98$_{\pm5.11}$ & 77.06$_{\pm4.48}$ 
                                                                       & &
                                                                       70.80$_{\pm2.64}$ &
                                                                       65.07$_{\pm3.75}$ &
                                                                       68.60$_{\pm4.81}$ 
                                                                       & &
                                                                       80.37$_{\pm5.34}$ &
                                                                       61.12$_{\pm7.17}$ &
                                                                       65.93$_{\pm6.65}$
                                                                       \\
                                                                       & \raggedright RRT-MIL~\cite{RRTMIL}  & 90.82$_{\pm3.45}$ & 84.27$_{\pm3.83}$  & 84.33$_{\pm3.82}$ &  & 97.42$_{\pm1.18}$ & 88.06$_{\pm3.37}$ & 89.86$_{\pm2.74}$ &  & 93.43$_{\pm1.21}$  & 72.92$_{\pm4.47}$ & 75.00$_{\pm4.81}$ 
                                                                       & &
                                                                       76.17$_{\pm4.62}$ &
                                                                       66.77$_{\pm4.29}$ &
                                                                       73.01$_{\pm3.49}$ 
                                                                       & &
                                                                       80.56$_{\pm4.78}$ &
                                                                       61.50$_{\pm5.66}$ &
                                                                       66.40$_{\pm5.26}$
                                                                       \\
                                                                       & \raggedright CoOp~\cite{coop}   & 87.36$_{\pm4.42}$  & 79.43$_{\pm5.30}$  & 79.57$_{\pm5.16}$ &  & 97.26$_{\pm0.64}$ & 88.19$_{\pm1.39}$ & 89.63$_{\pm1.66}$ &  &  92.98$_{\pm0.72}$  & 72.62$_{\pm2.89}$ & 75.54$_{\pm3.39}$ 
                                                                       & &
                                                                       76.03$_{\pm5.25}$ &
                                                                       66.94$_{\pm3.60}$ &
                                                                       70.17$_{\pm3.11}$ 
                                                                       & &
                                                                       81.22$_{\pm2.19}$ &
                                                                       61.56$_{\pm5.81}$ &
                                                                       67.55$_{\pm5.29}$
                                                                       \\
                                                                       & \raggedright CoCoOp~\cite{cocoop}  & 88.61$_{\pm2.71}$  & 80.26$_{\pm4.38}$  & 80.38$_{\pm4.23}$ &  & 97.36$_{\pm0.55}$ & 88.39$_{\pm0.91}$ & 89.94$_{\pm1.16}$ &  & 92.57$_{\pm1.94}$  & 72.10$_{\pm4.05}$ & 75.39$_{\pm3.94}$ 
                                                                       & &
                                                                       76.68$_{\pm3.67}$ &
                                                                       67.73$_{\pm3.16}$ &
                                                                       73.22$_{\pm3.21}$ 
                                                                       & &
                                                                       81.74$_{\pm2.16}$ &
                                                                       63.58$_{\pm2.97}$ &
                                                                       67.82$_{\pm2.39}$
                                                                       \\
                                                                       & \raggedright Metaprompt~\cite{metaprompt}  & 91.54$_{\pm3.39}$  & 83.74$_{\pm5.07}$  & 83.85$_{\pm4.91}$ &  & 
                                                                \uline{98.05$_{\pm0.32}$} & 90.47$_{\pm1.19}$ & \uline{91.81$_{\pm0.83}$} &  & 92.99$_{\pm2.03}$  & 74.94$_{\pm2.51}$ & 78.14$_{\pm2.73}$ 
                                                                       & &
                                                                       76.88$_{\pm2.02}$ &
                                                                \uline{68.02$_{\pm2.39}$} &
                                                                       72.95$_{\pm1.07}$ 
                                                                       & &
                                                                       82.67$_{\pm1.73}$ &
                                                                       65.21$_{\pm4.46}$ &
                                                                       69.17$_{\pm4.42}$
                                                                       \\
                                                                       & \raggedright TOP~\cite{qu2024rise} & 91.29$_{\pm 2.57}$      & 83.45$_{\pm 2.80}$    & 83.64$_{\pm 2.82}$    &  & 
                                                                       97.43$_{\pm 0.90}$      & 87.71$_{\pm 1.98}$    & 90.23$_{\pm 1.41}$  
                                                                       &  & 
                                                                       91.49$_{\pm 1.25}$      & 71.22$_{\pm 3.69}$    & 72.00$_{\pm 4.29}$ 
                                                                       & &
                                                                       76.48$_{\pm 2.41}$ &
                                                                       67.98$_{\pm 2.53}$ &
                                                                       71.22$_{\pm 2.85}$
                                                                       & &
                                                                       78.91$_{\pm4.62}$ &
                                                                       60.56$_{\pm4.38}$ &
                                                                       65.17$_{\pm3.86}$
                                                                       \\
                                                                       & \raggedright ViLa-MIL~\cite{shi2024vila}             & 90.97$_{\pm2.62}$  & 82.73$_{\pm2.97}$  & 82.81$_{\pm3.00}$ &  & 97.32$_{\pm0.85}$ & 86.21$_{\pm2.81}$ & 88.40$_{\pm2.57}$ &  & 92.91$_{\pm1.28}$  & 71.78$_{\pm3.93}$ & 73.58$_{\pm4.24}$   
                                                                       & &
                                                                       76.59$_{\pm1.83}$ &
                                                                       67.31$_{\pm1.98}$ &
                                                                       72.54$_{\pm1.96}$ 
                                                                       & &
                                                                       79.81$_{\pm5.01}$ &
                                                                       61.18$_{\pm5.46}$ &
                                                                       65.86$_{\pm4.79}$
                                                                       \\
                                                   & \raggedright \textbf{MSCPT(ours)} & \textbf{94.95$_{\pm1.47}$}  & \textbf{88.09$_{\pm2.32}$}  & \textbf{88.10$_{\pm2.32}$} &  & \textbf{98.55$_{\pm0.58}$} & \textbf{91.45$_{\pm2.16}$} & \textbf{92.95$_{\pm2.05}$} &  & \textbf{94.98$_{\pm0.60}$}  & \textbf{77.43$_{\pm2.11}$} & \textbf{79.57$_{\pm2.60}$}  
                                                   & &
                                                   \textbf{78.34$_{\pm2.52}$} &
                                                   \textbf{69.41$_{\pm2.54}$} &
                                                   \textbf{74.45$_{\pm3.04}$} 
                                                   & &
                                                   \textbf{85.39$_{\pm3.96}$} &
                                                   \textbf{66.42$_{\pm3.65}$} &
                                                   \textbf{71.56$_{\pm2.66}$}

\end{tblr}
\label{table2}
}
\vspace{-1.5em}
\end{table*}

\subsection{Comparisons with State-of-the-Art}
\label{iv-B}
The experimental results under the 16-shot setting are displayed in Table \ref{table1}. We observed some intriguing insights, such as complex and parameter-heavy methods like TransMIL and RRT-MIL underperformed despite their strong performance with full data training. Conversely, less parameterized methods such as ABMIL and CLAM exhibited slightly better performance. This is because traditional MIL-based methods require a large number of WSIs for training, and the more parameters they have, the more training data is needed. Furthermore, after adapting the prompt tuning methods designed for natural images (\textit{i.e.}, CoOp, CoCoOp, and Metaprompt) to tasks at the WSI-level, these methods outperform traditional MIL-based methods when based on CLIP and achieve comparable performance when using PLIP and CONCH. Relatively few parameters contribute to this result. Additionally, we found that Metaprompt outperforms CoOp across most metrics, thanks to its integration of visual prompt tuning and multi-scale information. This result motivates us to pursue visual prompt tuning and develop more effective multi-scale information integration modules. Despite prompt tuning methods designed for Few-shot Weakly Supervised WSI Classification (FSWC) tasks having a relatively higher number of parameters, they exhibit good performance. This leverages VLMs' prior knowledge through visual and text prompt tuning, reducing the need for extensive training samples.

Compared to other methods, MSCPT exhibits significant improvements in all evaluation metrics across the five datasets and three VLMs. Compared to the top traditional MIL-based methods, MSCPT shows the highest improvements in AUC, F1, and ACC by 13.0\%, 12.3\%, and 11.6\%, respectively. Overall, MSCPT shows greater performance improvements when based on CLIP compared to PLIP and CONCH. This is due to PLIP and CONCH being pre-trained on pathological images, which enhances representation ability and reduces reliance on textual descriptions. Compared to the top prompt tuning methods for natural images (\textit{i.e.}, CoOp, CoCoOp, and Metaprompt), MSCPT achieves the highest improvements of 7.6\%, 6.1\%, and 6.5\% in AUC, F1, and ACC, respectively.

\begin{table*}[t!]
\renewcommand{\arraystretch}{0.5}
\centering
\setlength{\extrarowheight}{0pt}
\setlength{\tabcolsep}{2.5pt}
\renewcommand{\arraystretch}{0.8}
\addtolength{\extrarowheight}{\aboverulesep}
\addtolength{\extrarowheight}{\belowrulesep}
\setlength{\aboverulesep}{0pt}
\setlength{\belowrulesep}{0pt}
\caption{Ablation experiment of five core components.}
\label{table3}
% \resizebox{0.48\textwidth}{!}{
\begin{tabular}{p{1cm}<{\centering} p{1cm}<{\centering} p{1cm}<{\centering} p{1cm}<{\centering} p{1cm}<{\centering} cccccccccccccccc} 
\toprule
\rule{0pt}{13pt}\multirow{2}{*}{\rotatebox{0}{\makecell{Prior\\Knowl.}}}  & \multirow{2}{*}{\rotatebox{0}{\makecell{Multi\\Scale}}}   &  \multirow{2}{*}{\rotatebox{0}{MHPT}}    & \multirow{2}{*}{\rotatebox{0}{ISGPT}} & \multirow{2}{*}{\rotatebox{0}{NPCGP}} & & \multicolumn{5}{c}{\makecell{\textbf{TCGA-NSCLC} \\ \scriptsize{(PLIP-based, 16-shot)}}}   &  & \multicolumn{5}{c}{\makecell{\textbf{PANDA} \\ \scriptsize{(CONCH-based, 4-shot)}}}  \\ 

\cline{7-11}
\cline{13-17}
 & &  &  &   & &   AUC   &  & F1     &  & ACC   &  &   AUC   &  & F1     &  & ACC  \\ 
\hline
    &    &    &   &   &  & 77.92$_{\pm5.48}$ &  & 71.58$_{\pm4.74}$  &  & 71.63$_{\pm4.75}$ & &  79.17$_{\pm2.91}$ &  & 56.21$_{\pm6.55}$  &  & 62.28$_{\pm6.25}$ \\
\checkmark  &   &   &   & & &  77.39$_{\pm5.92}$ &  & 71.47$_{\pm4.82}$  &  & 71.41$_{\pm4.87}$   & &  76.86$_{\pm4.72}$ &  & 57.71$_{\pm2.54}$  &  & 63.47$_{\pm3.67}$\\
    & \checkmark &   &   & &  & 78.31$_{\pm5.66}$ &  & 72.03$_{\pm4.60}$  &  & 71.86$_{\pm4.61}$  & & 81.48$_{\pm2.27}$ & & 59.96$_{\pm7.24}$ & & 65.97$_{\pm6.02}$\\
\checkmark   & \checkmark &   &   &   & & 78.53$_{\pm5.18}$ &  & 72.28$_{\pm4.52}$  &  & 72.02$_{\pm4.50}$ & & 80.46$_{\pm4.00}$ & &60.30$_{\pm3.27}$ & &65.57$_{\pm3.32}$ \\
\checkmark  & \checkmark &\checkmark &   &  &  &  80.57$_{\pm4.17}$ &  & 73.62$_{\pm2.41}$ &  & 73.77$_{\pm2.48}$    &  &  81.53$_{\pm3.03}$ &  & 61.86$_{\pm2.77}$ &  & 66.09$_{\pm3.01}$       \\
\checkmark  & \checkmark &\checkmark & \checkmark &   & &  82.75$_{\pm5.73}$  &  & 75.48$_{\pm4.70}$   &  & 75.53$_{\pm4.72}$  &  &  \textbf{82.51$_{\pm3.93}$} &  & 63.13$_{\pm1.95}$ &  & 67.22$_{\pm2.81}$ \\
\checkmark  & \checkmark  &\checkmark &   & \checkmark  &  & 81.92$_{\pm5.03}$          &  & 74.96$_{\pm4.68}$          &  & 75.10$_{\pm4.64}$  & &    81.90$_{\pm3.50}$ &  & 61.22$_{\pm1.93}$ &  & 66.54$_{\pm2.78}$       \\
\checkmark & & &\checkmark &\checkmark & &  79.44$_{\pm4.47}$  & & 73.06$_{\pm5.89}$ & & 72.59$_{\pm6.02}$ & &  82.06$_{\pm4.11}$ & & 60.33$_{\pm1.85}$ & &64.91$_{\pm3.00}$\\
\rowcolor[rgb]{0.93,0.93,0.93} \checkmark  & \checkmark  &\checkmark & \checkmark                      & \checkmark          &          &   \textbf{84.29$_{\pm3.97}$} &  & \textbf{76.39$_{\pm5.69}$} &  & \textbf{76.54$_{\pm5.49}$}  & &   82.33$_{\pm3.51}$  &  & \textbf{64.13$_{\pm1.51}$}          &  & \textbf{68.41$_{\pm2.46}$}  \\
\bottomrule
\end{tabular}
% }
\vspace{-1.5em}
\end{table*}

\begin{table}[tp]
\renewcommand{\arraystretch}{0.5}
\centering
\setlength{\extrarowheight}{0pt}
\setlength{\tabcolsep}{2.5pt}
\renewcommand{\arraystretch}{0.8}
\addtolength{\extrarowheight}{\aboverulesep}
\addtolength{\extrarowheight}{\belowrulesep}
\setlength{\aboverulesep}{0pt}
\setlength{\belowrulesep}{0pt}
\caption{Ablation experiment of Multi-scale Hierarchical Prompt Turning.}
\label{table4}
\resizebox{0.48\textwidth}{!}{
\begin{tabular}{p{1cm}<{\centering} p{1cm}<{\centering} p{1cm}<{\centering} ccccccccc} 
\toprule
\rule{0pt}{13pt} \multirow{2}{*}{Global}  & \multirow{2}{*}{High}   &  \multirow{2}{*}{Low} & & \multicolumn{5}{c}{\makecell[c]{\textbf{UBC-OCEAN}\\\scriptsize{(CONCH-based, 4-shot)}}}  \\ 

\cline{5-9}
 & & &  &   AUC   &  & F1     &  & ACC    \\ 
\hline
\checkmark  &  &   & &90.33$_{\pm2.46}$  &  &67.13$_{\pm5.55}$ &  &71.48$_{\pm5.40}$\\
\checkmark  & \checkmark  &   & &92.76$_{\pm1.92}$  &  &70.42$_{\pm4.85}$ &  &74.23$_{\pm4.86}$  \\
\checkmark  &  &\checkmark & &91.67$_{\pm2.29}$  &  &68.57$_{\pm5.12}$ &  &72.66$_{\pm5.03}$\\
\rowcolor[rgb]{0.93,0.93,0.93} \checkmark  & \checkmark  &\checkmark  &  & \textbf{93.19$_{\pm1.97}$} &  & \textbf{73.53$_{\pm5.26}$} &  & \textbf{76.81$_{\pm5.10}$} \\
\bottomrule
\end{tabular}
}
\vspace{-1.5em}
\end{table}

Prompt tuning methods explicitly designed for WSI exhibit superior performance on several datasets. This is attributed to their incorporation of priors into pre-trained VLMs and leveraging those priors to guide prompt tuning. Additionally, ViLa-MIL introduces multi-scale information compared to TOP, positioning it as the second-best overall performer. In comparison to ViLa-MIL, MSCPT shows improvements across all datasets, with AUC increasing by 0.5-5.3\%, F1 by 2.4-7.4\%, and ACC by 2.0-7.0\%. This is because MSCPT performs prompt tuning on both the textual and visual modalities. Furthermore, MSCPT takes into account both the multi-scale and contextual information of WSIs. Unlike the late fusion approach in ViLa-MIL, MSCPT employs an intermediate fusion method, leveraging the transformer layer and trainable global prompts to integrate pathological visual descriptions from both high and low levels. Moreover, compared to the brief prior knowledge incorporated in ViLa-MIL, MSCPT introduces more detailed pathological visual descriptions. This enables MSCPT to achieve greater performance improvements over ViLa-MIL as the VLM becomes more advanced.

\subsection{Results with Fewer Training Samples}
To further validate MSCPT's performance, we conducted experiments under 8-, 4-, and 2-shot settings. Notably, with limited training samples, sample selection critically influences model performance~\cite{qu2024rise}. However, this influence diminishes as the performance of the VLM improves. Accordingly, this experiment was conducted based on CONCH, with the results presented in Table \ref{table2}.

Overall, MSCPT consistently achieves the best performance across all datasets, with its relative advantage being most pronounced in low-data scenarios (\textit{e.g.}, 2-shot). Parameter-efficient methods such as CLAM and ABMIL demonstrate relatively stable performance, securing second-best results on certain datasets. In contrast, complex and parameter-heavy approaches like TransMIL and RRT-MIL perform poorly. Notably, MSCPT achieves significant improvements over the second-best model, with gains of up to 7.7\% in AUC, 8.1\% in F1, and 8.0\% in ACC.

\begin{table}[t]
\renewcommand{\arraystretch}{0.5}
\centering
\setlength{\extrarowheight}{0pt}
\setlength{\tabcolsep}{2pt}
\renewcommand{\arraystretch}{0.8}
\addtolength{\extrarowheight}{\aboverulesep}
\addtolength{\extrarowheight}{\belowrulesep}
\setlength{\aboverulesep}{0pt}
\setlength{\belowrulesep}{0pt}
\caption{Ablation experiment of different graph construction and training methods.}
\label{table5}
\resizebox{0.48\textwidth}{!}{
\begin{tabular}{lccccccc} 
\toprule
\rule{0pt}{13pt} \multirow{2}{*}{Methods}    & \multirow{2}{*}{\shortstack{Train.\\Param.}} &  & \multicolumn{5}{c}{\makecell[c]{\textbf{TCGA-RCC}\\\scriptsize{(CLIP-based, 16-shot)}}}  \\ 
\cline{4-8}
                                                          &                         &  & AUC                 &  & F1                  &  & ACC                  \\ 
\hline
MSCPT w/ TransMIL      & 2.44M    &  & 86.26$_{\pm4.74}$ &  & 68.54$_{\pm6.70}$  &  &    72.13$_{\pm6.87}$      \\
MSCPT w/ FA      & 2.45M          &  & 91.12$_{\pm4.14}$ &  & 71.40$_{\pm5.88}$  &  &       76.94$_{\pm5.13}$   \\
\hline
GCN w/ \textit{KNN(Coord.)}                                           & 1.35M                   &  & 92.85$_{\pm2.43}$~         &  & 79.29$_{\pm3.98}$          &  & 81.63$_{\pm3.74}$          \\
GCN w/ \textit{KNN(Feat.)}                                            & 1.35M                   &  & 93.92$_{\pm2.66}$~         &  & 80.46$_{\pm4.33}$          &  & 82.41$_{\pm4.26}$          \\
GAT w/ \textit{Sim.}                                                  & 1.35M                   &  & 93.14$_{\pm1.78}$~         &  & 80.82$_{\pm4.25}$          &  & 82.41$_{\pm3.85}$           \\
GraphSAGE w/ \textit{Sim.}                                            & 2.40M                   &  & 93.60$_{\pm2.72}$~         &  & 80.40$_{\pm3.89}$          &  & 82.66$_{\pm3.69}$           \\
\hline
\rowcolor[rgb]{0.93,0.93,0.93} \textbf{GCN w/ \textit{Sim.}(ours)} & \textbf{1.35M}                   &  & \textbf{95.04$_{\pm1.31}$~} &  & \textbf{82.59$_{\pm2.14}$} &  & \textbf{85.62$_{\pm2.14}$}  \\
\bottomrule
\end{tabular}
}
\vspace{-1.5em}
\end{table}

\begin{figure}[t]
   \begin{center}
   % \fbox{\rule{0pt}{2in} \rule{0.9\linewidth}{0pt}}ƒ
   \includegraphics[width=1\linewidth]{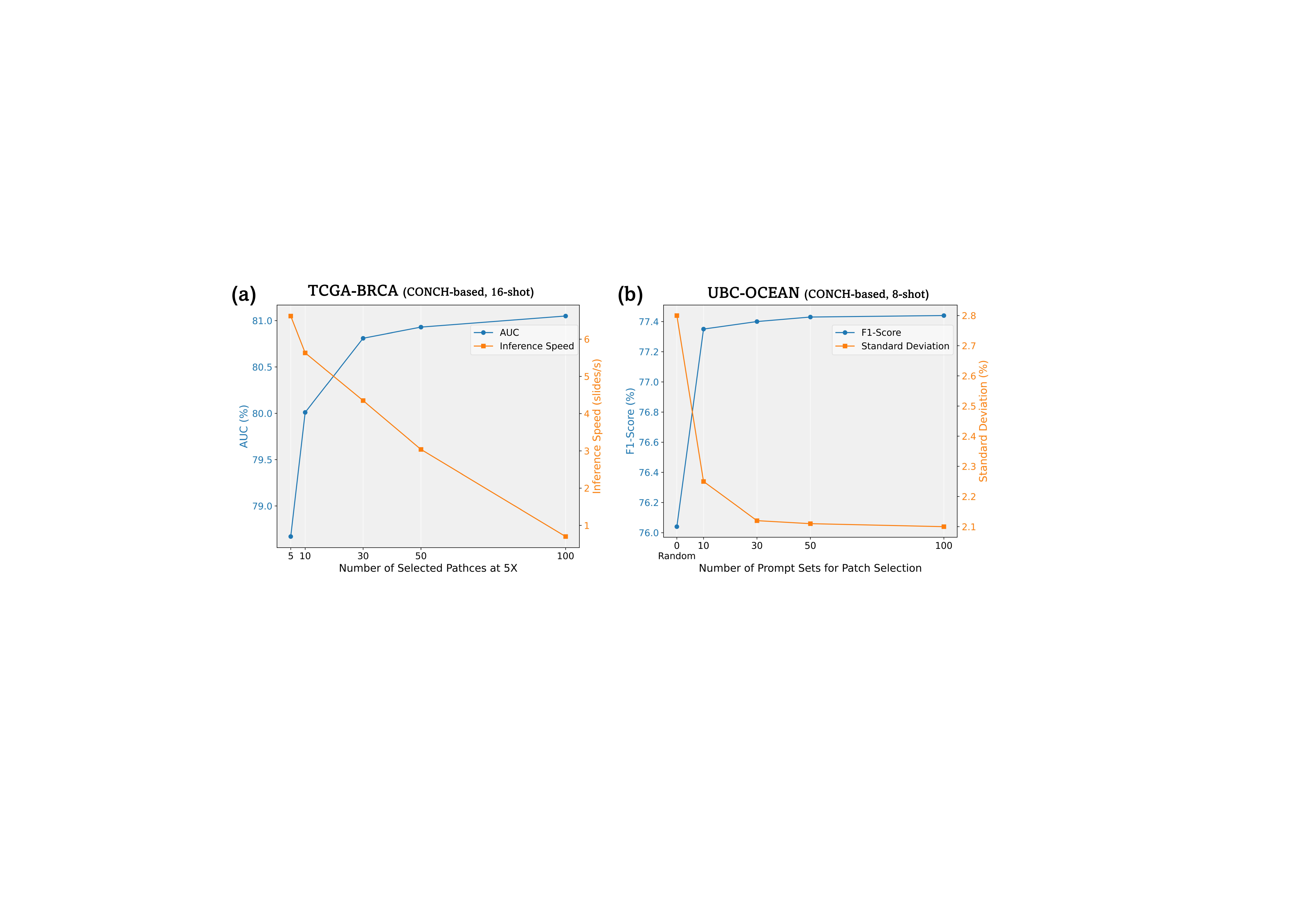}
   \end{center}
   \vspace{-0.5em}
   \caption{Ablation experiment of patch selection. \textbf{(a)} Influence of selected patch counts at 5× magnification in visual prompt tuning. \textbf{(b)} Influence of prompt set counts in patch selection.}
   \label{figure4}
   \vspace{-1em}
\end{figure}

Moreover, we observed some phenomena that differ from those in the 16-shot setting. Prompt-tuning methods explicitly designed for WSI experienced a significant drop in performance, particularly under the 2-shot setting. This decline can be attributed to both TOP and ViLa-MIL relying on CLIP, which struggles to comprehend complex and detailed pathological prior knowledge. As a result, these methods provide the VLM with only simplistic pathological priors. Such simplistic priors are challenging to fully exploit under extremely limited data conditions, leading to suboptimal performance. In contrast, MSCPT offers the VLM comprehensive and detailed pathological visual priors across multiple scales, enabling it to leverage all available information, even when data is scarce. Metaprompt also demonstrates superior performance under limited training data, thanks to its ability to capture multi-scale information and utilize visual prompt tuning. Our MSCPT effectively integrates these advantages, resulting in robust performance.

\subsection{Ablation Experiment}
\subsubsection{Effects of Each Component in MSCPT}
To verify the effectiveness of the five core components, ablation experiments were conducted on the TCGA-NSCLC and PANDA datasets. The experimental results are presented in Table \ref{table3}. When all modules were removed, MSCPT regressed to CoOp~\cite{coop}. Integrating a multi-scale framework with visual prompt tuning into CoOp aligns its implementation with that of Metaprompt. The incorporation of visual prompt tuning and multi-scale information enhances the model’s performance, particularly in scenarios with limited training samples. While CoOp and Metaprompt use class names as text prompts, we introduce more detailed visual prior knowledge as their new text prompts. However, experimental results demonstrate that the performance of CoOp and Metaprompt did not consistently improve after incorporating visual prior knowledge; in fact, there was even a slight decrease in performance. We argue that the prompt tuning methods used in CoOp and Metaprompt, which are designed for natural images, fail to fully leverage the structure of prior knowledge. Furthermore, the use of excessively long text prompts negatively impacts the results.

All metrics improved by 0.8-2.9\% after integrating the Multi-scale Hierarchical Prompt Tuning (MHPT) module into Metaprompt. This is because the MHPT module utilizes transformer layers to integrate pathological visual descriptions across different scales, enhances the model's information aggregation capabilities. Unlike CoOp, Metaprompt, and ViLa-MIL, which can only incorporate a single, relatively long visual prior knowledge prompt, MSCPT inputs a large amount of detailed visual prior knowledge in separate fragments. This enhances MSCPT's ability to comprehend and leverage prior knowledge more effectively. Building upon this, we introduced the Image-text Similarity-based Graph Prompt Tuning (ISGPT) module, which led to improvements in all metrics (1.2-2.7\%). This demonstrated that utilizing ISGPT for contextual learning also enhances model performance, reaffirming the importance of contextual information for WSI analysis. When we added both the MHPT and Non-Parametric Cross-Guided Pooling (NPCGP) module to the baseline, in comparison to solely adding MHPT, the metrics improved by at most 1.8\%. This indicates that the NPCGP module, compared to attention-based pooling, is more effective in identifying important patches within the WSI, resulting in better instance aggregation results. Additionally, NPCGP uses image-text similarity for classification, offering more human-interpretable results compared to attention-based methods.

Furthermore, the importance of the multi-scale architecture in MSCPT was validated by removing the low-level encoders. Without low-level descriptions, the MHPT module retained only global and high-level prompts, while ISGPT and NPCGP were simplified to single-scale structures. As shown in Table \ref{table3}, this resulted in 4.4\% and 6.0\% decreases in F1-scores on the TCGA-NSCLC and PANDA datasets, respectively. These results show that the multi-scale structure, despite higher computational complexity, significantly improves performance.

\subsubsection{Effects of Patch Selection}
Unlike previous prompt tuning methods designed for WSIs, MSCPT incorporates visual prompt tuning. However, given the enormous size of WSIs, it is impractical to perform prompt tuning on all patches. To address this, MSCPT leverages the zero-shot capability of VLMs to preselect high-value patches for subsequent prompt tuning. As shown in Fig. \ref{figure4}\textcolor{red}{a}, we conducted an ablation study on the number of patches used for visual prompt tuning on the TCGA-BRCA dataset. The results show that increasing the number of patches for visual prompt tuning improves AUC but significantly increases inference time and GPU memory usage. Therefore, we adopt a trade-off by selecting 30 patches per class, which achieves strong performance while maintaining relatively fast inference time.

Similarly, an ablation study was conducted on the number of prompt sets used for patch selection on the UBC-OCEAN dataset (Fig. \ref{figure4}\textcolor{red}{b}). When patches were randomly selected, the model's performance showed a significant drop, accompanied by an increase in standard deviation. In contrast, increasing the number of prompt sets for patch selection improved the model's robustness. Moreover, the number of prompt sets had a minimal impact on the results. Considering these factors, we selected 50 prompt sets for patch selection.

\begin{figure*}[t!]
   \begin{center}
   % \fbox{\rule{0pt}{2in} \rule{0.9\linewidth}{0pt}}ƒ
   \includegraphics[width=1\linewidth]{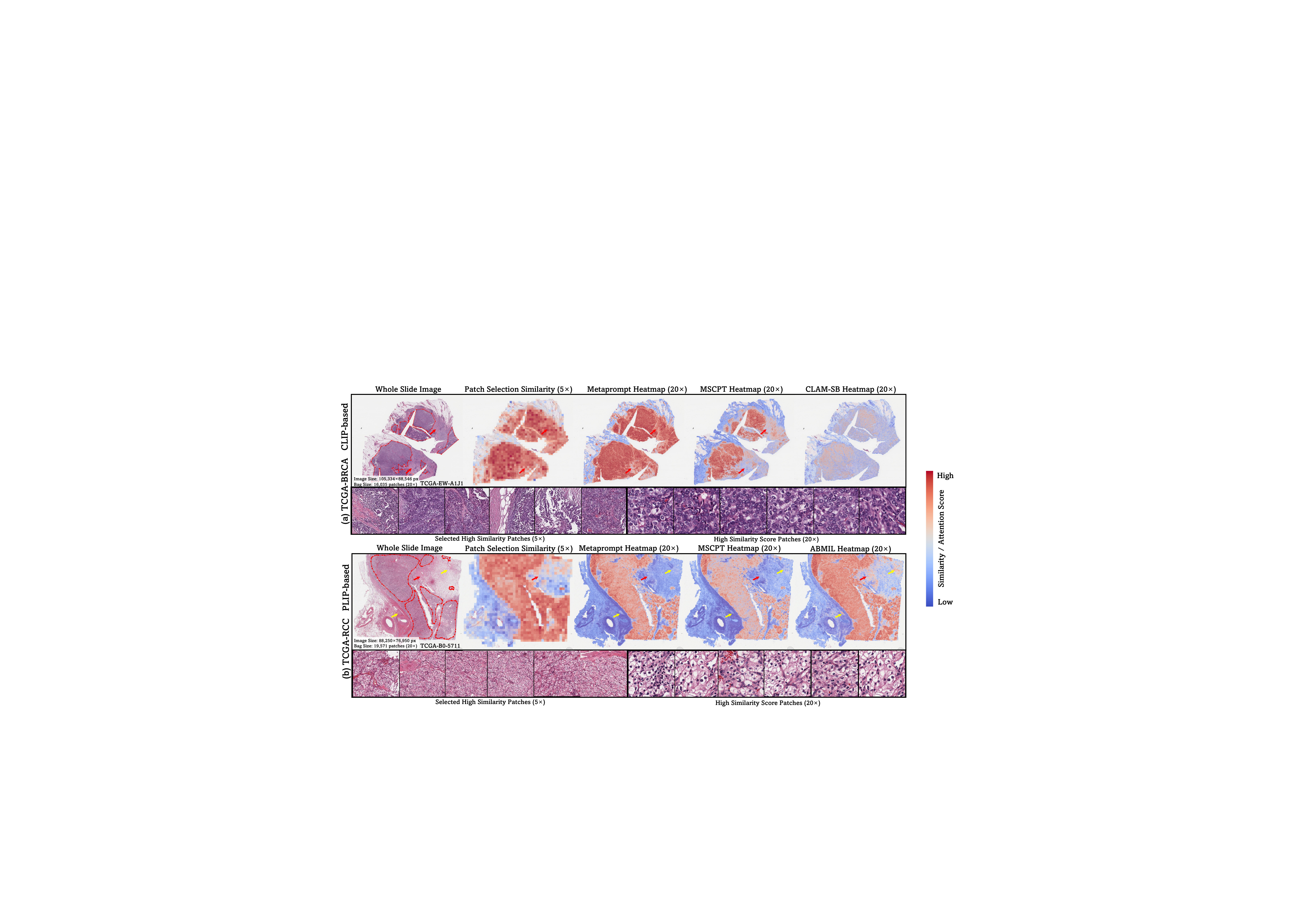}
   \end{center}
   \vspace{-1em}
   \caption{Visualization of the original WSI, similarity score map for patch selection, heatmaps generated using MSCPT, attention heatmaps of Metaprompt and the top traditional method, selected high similarity patches and patches with highest similarity scores. The area surrounded by the \textcolor[RGB]{255,0,0}{red line} in the WSI indicates the tumor area. The similarity score is derived from the image-text similarity of the NPCGP module.}
   \label{figure5}
   \vspace{-1em}
\end{figure*}

\subsubsection{Effects of Different Prompts in MHPT}
We conduct an ablation study on base-to-new generalization using various prompt combinations in Hierarchical Prompt Tuning (MHPT), with the results presented in Table \ref{table4}. The baseline method utilizes only global prompts. To eliminate the influence of high-level text prompts, we replace the high-level visual descriptions with meaningless placeholders. Similarly, the impact of low-level text prompts is analyzed by removing the prompt generator and excluding low-level text prompts from the MHPT framework. The experimental results demonstrate that both high-level and low-level text prompts positively contribute to the model’s performance. Adding high-level text prompts results in the most significant performance improvement compared to the baseline. Unlike providing pathological prior knowledge to CoOp and Metaprompt, our subsequent ISGPT module fully leverages detailed, fragment-based representations of prior knowledge to enhance performance. When the prompt generator integrates low-level visual descriptions into the prompted text encoder, the model performance further improves. Notably, even with a frozen text encoder, the presence of the prompt generator enables MSCPT to better adapt to low-level visual descriptions.

\subsubsection{Effects of Image-text Similarity-based Graph Prompt Tuning}
MSCPT utilizes image-text similarity to build a graph neural network that effectively captures the global information in WSIs. To assess the efficacy of the Image-text Similarity-based Graph Prompt Tuning (ISGPT) module, we conducted a comprehensive ablation study. Experimental results on TCGA-RCC are summarized in Table \ref{table5}.

First, we replaced the ISGPT module with a Transformer, which, in theory, can dynamically learn the contextual relationships between image embeddings and text embeddings. Specifically, we concatenated the dual-scale image and text embeddings obtained after four encoding layers and fed them into a Self-Attention module to dynamically capture the global information of WSIs. However, due to the enormous size of pathology images, directly applying the native Self-Attention mechanism is impractical. To address this, we utilized two alternative approaches: a modified version of TransMIL~\cite{shao2021transmil} (excluding the prediction head) and FlashAttention (FA)~\cite{dao2023flashattention2} to perform the experiments. However, experiments show that replacing the ISGPT module with a Transformer resulted in a significant performance drop. This is due to the reparameterization nature of the Transformer, which prevents it from being sufficiently trained in a few-shot setting.

To validate the effectiveness of constructing adjacency matrices via image-text similarity (\textit{Sim.}), we compared it with k-nearest neighbors (KNN) using patch coordinates (\textit{KNN(Coord.)}) or visual features (\textit{KNN(Feat.)}), as in prior studies~\cite{patchgcn, h2gt}. Additionally, we tested the effectiveness of GCN by comparing them with GAT~\cite{gat} and GraphSAGE~\cite{graphsage}.  Switching to KNN to construct the adjacency matrix led to decreased performance across all metrics, whether based on coordinates or patch visual features. This decline may be attributed to the limited scope of connectivity in the adjacent matrix construction methods. Connecting only nearby patches based on coordinates restricts the GNN to the local context, while connecting visually similar patches based on their features may lack global information about interactions between different types of tissue organization. In contrast, our ISGPT module connects patches related to specific cancer types, overcoming local or visual similarity connection limitations and enabling a more comprehensive contextual understanding. When we replaced GCN with GAT or GraphSAGE, the model's performance also experienced varying degrees of decline. We demonstrate that complex graph neural networks are unsuitable for few-shot scenarios.

\begin{table}[t]
\renewcommand{\arraystretch}{0.5}
\centering
\setlength{\extrarowheight}{0pt}
\setlength{\tabcolsep}{2.5pt}
\renewcommand{\arraystretch}{0.8}
\addtolength{\extrarowheight}{\aboverulesep}
\addtolength{\extrarowheight}{\belowrulesep}
\setlength{\aboverulesep}{0pt}
\setlength{\belowrulesep}{0pt}
\caption{ Ablation experiment of different instance aggregation methods.}
\label{table6}
\begin{tabular}{lcccc} 
\toprule
\rule{0pt}{13pt}\multirow{2}{*}{Methods} & \multirow{2}{*}{} & \multicolumn{3}{c}{\makecell[c]{\textbf{TCGA-NSCLC}\\\scriptsize{(PLIP-based, 16-shot)}}}\\ 
\cline{3-5}
                                                             &                   & AUC                 & F1                  & ACC                  \\ 
\hline
Mean Pooling                                                 &                   & 78.88$_{\pm4.61}$          & 73.60$_{\pm3.38}$         & 73.39$_{\pm3.68}$           \\
Max Pooling                                                  &                   & 82.63$_{\pm4.58}$          & 75.68$_{\pm3.51}$          & 75.89$_{\pm3.54}$           \\
Attention-based Pooling                                      &                   & 82.75$_{\pm5.73}$          & 75.48$_{\pm4.70}$          & 75.53$_{\pm4.72}$           \\
\rowcolor[rgb]{0.93,0.93,0.93}NPCGP w/o cross-guidance          &                   & 83.61$_{\pm5.68}$          & 75.98$_{\pm5.35}$          & 75.81$_{\pm5.21}$           \\
\rowcolor[rgb]{0.93,0.93,0.93} \textbf{NPCGP(ours)} &                               & \textbf{84.29$_{\pm3.97}$} & \textbf{76.39$_{\pm5.69}$} & \textbf{76.54$_{\pm5.49}$}  \\
\bottomrule
\end{tabular}
\vspace{-1.5em}
\end{table}

\subsubsection{Effects of Instance Aggregation}
\begin{figure*}[t!]
   \begin{center}
   % \fbox{\rule{0pt}{2in} \rule{0.9\linewidth}{0pt}}ƒ
   \includegraphics[width=1\linewidth]{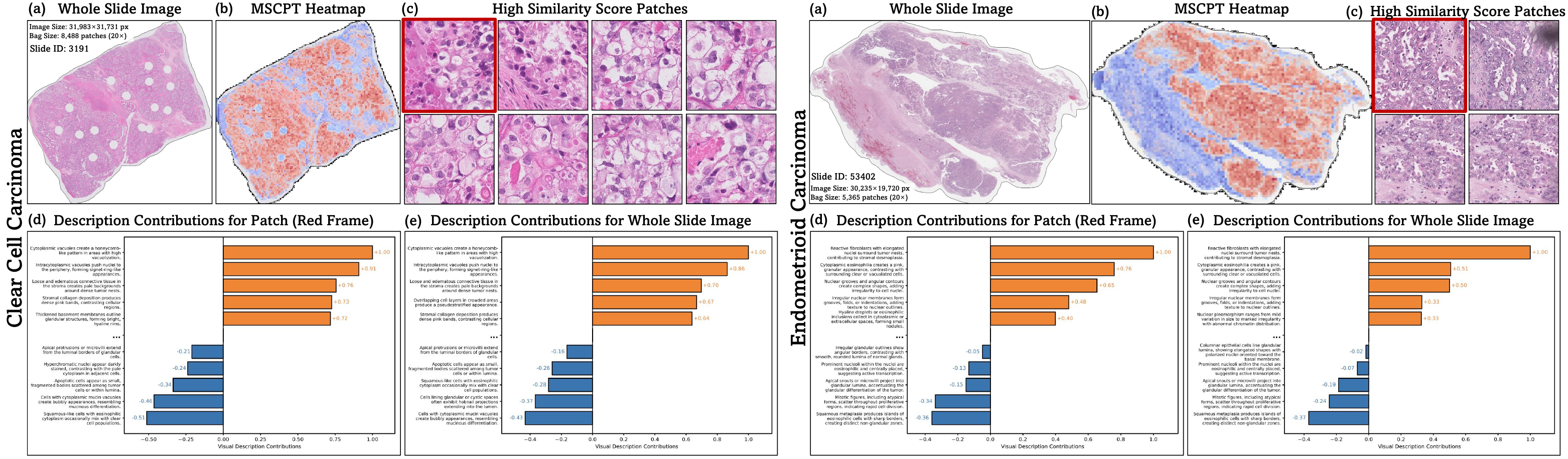}
   \end{center}
   \vspace{-1em}
   \caption{Interpretability analysis on the UBC-OCEAN dataset. Including: \textbf{(a)} original WSI, \textbf{(b)} heatmap generated by MSCPT, \textbf{(c)} selected patches with the highest similarity score, \textbf{(d)} description contributions for the selected patch, and \textbf{(e)} description contributions for WSI.}
   \label{figure6}
   \vspace{-1em}
\end{figure*}

To validate the effectiveness of our instance aggregation method, we compared Non-Parametric Cross-Guided Pooling (NPCGP) with other aggregation methods (\textit{i.e.}, Mean, Max, and Attention-based Pooling). The results based on PLIP on the TCGA-NSCLC are presented in Table \ref{table6}. When we replaced NPCGP with other methods, the performance of the models decreased to varying degrees. This implies that our NPCGP can discern more impactful patches and aggregate them into bag features. The visualization results in Section \ref{visualization} support this point. We also conducted an ablation study on cross-guidance. Instead of computing cross-scale cosine similarity during feature aggregation, we only calculated cosine similarity between image and text embeddings at the same scale. Removing cross-guidance led to a drop in performance across all metrics, likely because LLMs may produce descriptions with incorrect scales. In addition, NPCGP utilizes image-text similarity to classify each WSI, which enhances the interpretability of MSCPT (see Fig. \ref{figure6} and Section \ref{interpretability}).

\begin{table}[t]
\renewcommand{\arraystretch}{0.5}
\centering
\setlength{\extrarowheight}{0pt}
\setlength{\tabcolsep}{4pt}
\renewcommand{\arraystretch}{0.8}
\addtolength{\extrarowheight}{\aboverulesep}
\addtolength{\extrarowheight}{\belowrulesep}
\setlength{\aboverulesep}{0pt}
\setlength{\belowrulesep}{0pt}
\caption{Ablation experiment of different large language models.}
\label{table7}
\begin{tabular}{lcccc} 
\toprule
\rule{0pt}{13pt} \multirow{2}{*}{Methods}   & \multirow{2}{*}{} & \multicolumn{3}{c}{\makecell[c]{\textbf{TCGA-RCC}\\\scriptsize{(CLIP-based, 16-shot)}}}\\ 
\cline{3-5}
                                        &                   & AUC        & F1         & ACC         \\ 
\cline{1-5}
Gemini-1.5-pro~\cite{team2023gemini}                                &                   & 94.14$_{\pm1.79}$ & 81.61$_{\pm2.69}$ & 83.67$_{\pm3.63}$  \\
Claude-3~\cite{claude3}                                &                   & 93.97$_{\pm2.35}$ & 80.61$_{\pm4.05}$ & 82.72$_{\pm3.59}$  \\
Llama-3~\cite{llama}                                 &                   & 94.63$_{\pm1.42}$ & 82.36$_{\pm1.81}$ & 84.38$_{\pm2.08}$  \\
GPT-3.5~\cite{gpt3} &                   & 94.52$_{\pm2.16}$ & 82.09$_{\pm3.07}$ & 84.27$_{\pm3.61}$  \\
\rowcolor[rgb]{0.93,0.93,0.93} \textbf{GPT-4}~\cite{gpt4} &                   & \textbf{95.04$_{\pm1.31}$} & \textbf{83.78$_{\pm2.19}$} & \textbf{85.62$_{\pm2.14}$}  \\
\bottomrule
\end{tabular}
\vspace{-1.5em}
\end{table}

\subsubsection{Effects of Large Language Models}
To verify the impact of different LLMs on model performance, we compared the performance of MSCPT when using descriptions generated by different LLMs (\textit{i.e.}, Gemini-1.5-pro~\cite{team2023gemini}, Claude-3~\cite{claude3}, Llama-3~\cite{llama}, GPT-3.5~\cite{gpt3}, and GPT-4~\cite{gpt4}). The results obtained using CLIP on TCGA-RCC are presented in Table \ref{table7}. When generating descriptions using Claude-3, MSCPT performs comparably to the baseline. However, MSCPT outperforms the baseline when using other LLMs. This demonstrates MSCPT's robustness across different LLMs and underscores the importance of accurate visual descriptions in enhancing model performance.

\subsection{Visualization Analysis}
\label{visualization}
We performed visualizations of MSCPT, Metaprompt, and the best-performing traditional MIL-based method on one case of PLIP-based TCGA-RCC and one case of CLIP-based TCGA-BRCA. As depicted in Fig. \ref{figure5}\textcolor{red}{a}, during patch selection, CLIP assigned high similarity scores to tumor regions as expected, but unexpectedly also to non-tumor areas. This outcome arose because CLIP was not specifically designed for pathological images, resulting in a less-than-optimal zero-shot capability for this type of imagery. However, after prompt tuning using MSCPT, the model correctly assigned high scores to the actual tumor regions, while the regions that originally received high scores dropped to lower score ranges (red arrows). Meanwhile, CLAM-SB struggled to differentiate tumors and non-tumors. Similarly, Metaprompt assigned high attention weights to certain non-tumor tissues (red arrows).

When selecting patches using PLIP, the model could roughly identify tumor regions but also assigned high confidence scores to some non-tumor areas (Fig. \ref{figure5}\textcolor{red}{b}). This limitation was effectively addressed by both MSCPT and Metaprompt, as indicated by the red arrows. While ABMIL could also determine instance importance, it tended to assign higher scores to certain non-tumor regions compared to MSCPT and Metaprompt (yellow arrows). Due to PLIP's improved ability to represent pathological images, Metaprompt produced visualization results comparable to MSCPT. The patches with high similarity scores exhibit features such as enlarged nuclei, increased nuclear density, and irregularity, which are key characteristics of cancer.

\subsection{Interpretability Analysis}
\label{interpretability}
The probability distribution output by MSCPT is derived from the similarity between the pathological image and the pathological visual description. Therefore, this similarity can be regarded as an indicator of the contribution to the final classification decision. The more an image resembles a specific pathological visual description for a given category, the more likely MSCPT is to classify the WSI as that category. We visualize the similarity between a given pathological image and all pathological visual descriptions within that category, which allows us to assess the contribution of each description to the image's classification. Similarly, by averaging the similarities selected using Top-K, we can assess each description's contribution to the overall WSI classification.

As shown in Fig. \ref{figure6}, we performed an interpretability analysis on one case each from the Clear Cell Carcinoma and Endometrioid Carcinoma of the UBC-OCEAN dataset. Panels (d) and (e) show the contribution of pathological visual descriptions to the selected patch and WSI at high resolution, highlighting only the top five and bottom five descriptions. For the selected patch in Clear Cell Carcinoma, the two most contributing pathological visual descriptions are: \textit{``Cytoplasmic vacuoles create a honeycomblike pattern in areas with highvacuolization"} and \textit{``Intracytoplasmic vacuoles push nuclei to the periphery, forming signet-ring-like appearances"}. The descriptions corresponding to the selected image from Endometrioid Carcinoma are: \textit{``Reactive fibroblasts with elongated nuclei surround tumor nests, contributing to stromal desmoplasia"} and \textit{``Cytoplasmic eosinophilia creates a pink, granular appearance, contrasting with surrounding clear or vacuolated cells"}. These pathological visual descriptions closely match the features and patterns observed in the image, demonstrating a high degree of consistency with its content.

\section{Conclusion and Discussion}
\vspace{0.3pt}\noindent\textbf{Conclusion.}\hspace{1ex}
In this paper, we propose \textbf{M}ulti-\textbf{S}cale and \textbf{C}ontext-focused \textbf{P}rompt \textbf{T}uning (MSCPT) to address the Few-shot Weakly-supervised WSI Classification (FSWC) task. MSCPT leverages GPT-4~\cite{gpt4} to generate multi-scale pathological visual descriptions, which guide hierarchical prompt tuning and instance aggregation. Experimental results on five WSI datasets, three downstream tasks, and three Vision-Language Models (VLMs) demonstrate that MSCPT achieves state-of-the-art performance in FSWC tasks. Additionally, MSCPT enables effective fine-tuning of diverse VLMs for WSI-level tasks while offering significant advantages in interpretability compared to traditional MIL-based methods.

\vspace{0.3pt}\noindent\textbf{Limitations.}\hspace{1ex}
However, we observe substantial variations in performance across different datasets and VLMs, primarily due to the dependence of fine-tuning effectiveness on the quality and suitability of pre-trained VLMs. With the rapid advancements in computational pathology, increasingly comprehensive and powerful pre-trained pathological VLMs are emerging, greatly facilitating the resolution of FSWC tasks. We believe that MSCPT, when paired with these advanced VLMs, will serve as a robust and versatile tool in FSWC task and computational pathology.

\vspace{0.3pt}\noindent\textbf{Future Work.}\hspace{1ex}
We outline three key strategic directions for future research: 1) \textit{Multimodal Large Language Models}: More powerful multimodal large language models (MLLMs) specifically designed for pathology are rapidly advancing~\cite{sun2024cpath,sun2024pathgen}. We plan to extend the idea of MSCPT to MLLMs to achieve better performance; 2) \textit{Multi-Modal and Multi-Omics Integration}: Integrating WSI-derived visual patterns with genomic data is a promising direction. By creating a unified framework for combined histomorphological-molecular classification, we strive to unlock deeper insights into disease mechanisms and improve diagnostic precision; 3) \textit{Ultra-Data-Efficient Learning}: Extending few-shot capabilities to one/zero-shot scenarios via iterative prompt refinement and self-supervised pretext tasks. Advancing developments in these areas will help bridge the gap between AI and clinical practice, ultimately fostering more interpretable and better performing diagnostic outcomes.

\bibliographystyle{IEEEtran}
\bibliography{reference}

% Generated by IEEEtran.bst, version: 1.14 (2015/08/26)
\begin{thebibliography}{10}
\providecommand{\url}[1]{#1}
\csname url@samestyle\endcsname
\providecommand{\newblock}{\relax}
\providecommand{\bibinfo}[2]{#2}
\providecommand{\BIBentrySTDinterwordspacing}{\spaceskip=0pt\relax}
\providecommand{\BIBentryALTinterwordstretchfactor}{4}
\providecommand{\BIBentryALTinterwordspacing}{\spaceskip=\fontdimen2\font plus
\BIBentryALTinterwordstretchfactor\fontdimen3\font minus \fontdimen4\font\relax}
\providecommand{\BIBforeignlanguage}[2]{{%
\expandafter\ifx\csname l@#1\endcsname\relax
\typeout{** WARNING: IEEEtran.bst: No hyphenation pattern has been}%
\typeout{** loaded for the language `#1'. Using the pattern for}%
\typeout{** the default language instead.}%
\else
\language=\csname l@#1\endcsname
\fi
#2}}
\providecommand{\BIBdecl}{\relax}
\BIBdecl

\bibitem{shao2021transmil}
Z.~Shao, H.~Bian \emph{et~al.}, ``Transmil: Transformer based correlated multiple instance learning for whole slide image classification,'' \emph{NeurIPS}, vol.~34, pp. 2136--2147, 2021.

\bibitem{xing2024comprehensive}
X.~Xing, M.~Zhu \emph{et~al.}, ``Comprehensive learning and adaptive teaching: Distilling multi-modal knowledge for pathological glioma grading,'' \emph{Med. Image Anal.}, vol.~91, p. 102990, 2024.

\bibitem{guo2023predicting}
Q.~Guo, L.~Qu \emph{et~al.}, ``Predicting lymph node metastasis from primary cervical squamous cell carcinoma based on deep learning in histopathologic images,'' \emph{Mod. Pathol.}, vol.~36, no.~12, p. 100316, 2023.

\bibitem{ludwig2005biomarkers}
J.~A. Ludwig and J.~N. Weinstein, ``Biomarkers in cancer staging, prognosis and treatment selection,'' \emph{Nat. Rev. Cancer}, vol.~5, no.~11, pp. 845--856, 2005.

\bibitem{ilse2020deep}
M.~Ilse, J.~M. Tomczak \emph{et~al.}, ``Deep multiple instance learning for digital histopathology,'' in \emph{MICCAI}.\hskip 1em plus 0.5em minus 0.4em\relax Elsevier, 2020, pp. 521--546.

\bibitem{qu2024rise}
L.~Qu, K.~Fu \emph{et~al.}, ``The rise of ai language pathologists: Exploring two-level prompt learning for few-shot weakly-supervised whole slide image classification,'' \emph{NeurIPS}, vol.~36, 2024.

\bibitem{shao2023hvtsurv}
Z.~Shao, Y.~Chen \emph{et~al.}, ``Hvtsurv: Hierarchical vision transformer for patient-level survival prediction from whole slide image,'' in \emph{AAAI}, vol.~37, no.~2, 2023, pp. 2209--2217.

\bibitem{campanella2019clinical}
G.~Campanella, M.~G. Hanna \emph{et~al.}, ``Clinical-grade computational pathology using weakly supervised deep learning on whole slide images,'' \emph{Nat. Med.}, vol.~25, no.~8, pp. 1301--1309, 2019.

\bibitem{qu2024rethinking}
L.~Qu, Y.~Ma \emph{et~al.}, ``Rethinking multiple instance learning for whole slide image classification: A good instance classifier is all you need,'' \emph{IEEE Trans. Circuits Syst. Video Technol.}, 2024.

\bibitem{srinidhi2021deep}
C.~L. Srinidhi, O.~Ciga \emph{et~al.}, ``Deep neural network models for computational histopathology: A survey,'' \emph{Med. Image Anal.}, vol.~67, p. 101813, 2021.

\bibitem{shmatko2022artificial}
A.~Shmatko, N.~Ghaffari~Laleh \emph{et~al.}, ``Artificial intelligence in histopathology: enhancing cancer research and clinical oncology,'' \emph{Nat. Cancer}, vol.~3, no.~9, pp. 1026--1038, 2022.

\bibitem{clip}
A.~Radford, J.~W. Kim \emph{et~al.}, ``Learning transferable visual models from natural language supervision,'' in \emph{ICML}.\hskip 1em plus 0.5em minus 0.4em\relax PMLR, 2021, pp. 8748--8763.

\bibitem{lee2020biobert}
J.~Lee, W.~Yoon \emph{et~al.}, ``Biobert: a pre-trained biomedical language representation model for biomedical text mining,'' \emph{Bioinformatics}, vol.~36, no.~4, pp. 1234--1240, 2020.

\bibitem{maple}
M.~U. Khattak, H.~Rasheed \emph{et~al.}, ``Maple: Multi-modal prompt learning,'' in \emph{CVPR}, 2023, pp. 19\,113--19\,122.

\bibitem{blip}
J.~Li, D.~Li \emph{et~al.}, ``Blip: Bootstrapping language-image pre-training for unified vision-language understanding and generation,'' in \emph{ICML}.\hskip 1em plus 0.5em minus 0.4em\relax PMLR, 2022, pp. 12\,888--12\,900.

\bibitem{GexMolGen}
J.~Cheng, X.~Pan \emph{et~al.}, ``Gexmolgen: cross-modal generation of hit-like molecules via large language model encoding of gene expression signatures,'' \emph{Brief. Bioinform.}, vol.~25, no.~6, p. bbae525, 2024.

\bibitem{han2024umpire}
M.~Han, D.~Yang \emph{et~al.}, ``Towards unified molecule-enhanced pathology image representation learning via integrating spatial transcriptomics,'' \emph{arXiv preprint arXiv:2412.00651}, 2024.

\bibitem{mizero}
M.~Y. Lu, B.~Chen \emph{et~al.}, ``Visual language pretrained multiple instance zero-shot transfer for histopathology images,'' in \emph{CVPR}, 2023, pp. 19\,764--19\,775.

\bibitem{plip}
Z.~Huang, F.~Bianchi \emph{et~al.}, ``A visual--language foundation model for pathology image analysis using medical twitter,'' \emph{Nat. Med.}, vol.~29, no.~9, pp. 2307--2316, 2023.

\bibitem{conch}
M.~Y. Lu, B.~Chen \emph{et~al.}, ``A visual-language foundation model for computational pathology,'' \emph{Nat. Med.}, vol.~30, p. 863–874, 2024.

\bibitem{cocoop}
K.~Zhou, J.~Yang \emph{et~al.}, ``Conditional prompt learning for vision-language models,'' in \emph{CVPR}, 2022, pp. 16\,816--16\,825.

\bibitem{coop}
{Zhou, Kaiyang and Yang, Jingkang and Loy, Chen Change and Liu, Ziwei}, ``Learning to prompt for vision-language models,'' \emph{Int. J. Comput. Vis.}, vol. 130, no.~9, pp. 2337--2348, 2022.

\bibitem{metaprompt}
C.~Zhao, Y.~Wang \emph{et~al.}, ``Learning domain invariant prompt for vision-language models,'' \emph{IEEE Trans. Image Process.}, 2024.

\bibitem{chen2022scaling}
R.~J. Chen, C.~Chen \emph{et~al.}, ``Scaling vision transformers to gigapixel images via hierarchical self-supervised learning,'' in \emph{CVPR}, June 2022, pp. 16\,144--16\,155.

\bibitem{shao2023characterizing}
W.~Shao, Y.~Zuo \emph{et~al.}, ``Characterizing the survival-associated interactions between tumor-infiltrating lymphocytes and tumors from pathological images and multi-omics data,'' \emph{IEEE Trans. Med. Imaging}, 2023.

\bibitem{patchgcn}
R.~J. Chen, M.~Y. Lu \emph{et~al.}, ``Whole slide images are 2d point clouds: Context-aware survival prediction using patch-based graph convolutional networks,'' in \emph{MICCAI}.\hskip 1em plus 0.5em minus 0.4em\relax Springer International Publishing, 2021, pp. 339--349.

\bibitem{h2gt}
M.~Han, X.~Zhang \emph{et~al.}, ``Multi-scale heterogeneity-aware hypergraph representation for histopathology whole slide images,'' in \emph{ICME}.\hskip 1em plus 0.5em minus 0.4em\relax IEEE, 2024, pp. 1--6.

\bibitem{clam}
M.~Y. Lu, D.~F. Williamson \emph{et~al.}, ``Data-efficient and weakly supervised computational pathology on whole-slide images,'' \emph{Nat. Biomed. Eng}, vol.~5, no.~6, pp. 555--570, 2021.

\bibitem{abmil}
M.~Ilse, J.~Tomczak \emph{et~al.}, ``Attention-based deep multiple instance learning,'' in \emph{ICML}.\hskip 1em plus 0.5em minus 0.4em\relax PMLR, 2018, pp. 2127--2136.

\bibitem{dsmil}
B.~Li, Y.~Li \emph{et~al.}, ``Dual-stream multiple instance learning network for whole slide image classification with self-supervised contrastive learning,'' in \emph{CVPR}, 2021, pp. 14\,318--14\,328.

\bibitem{gtmil}
Y.~Zheng, R.~H. Gindra \emph{et~al.}, ``A graph-transformer for whole slide image classification,'' \emph{IEEE Trans. Med. Imaging}, vol.~41, no.~11, pp. 3003--3015, 2022.

\bibitem{schuhmann2022laion}
C.~Schuhmann, R.~Beaumont \emph{et~al.}, ``Laion-5b: An open large-scale dataset for training next generation image-text models,'' \emph{NeurIPS}, vol.~35, pp. 25\,278--25\,294, 2022.

\bibitem{shi2024vila}
J.~Shi, C.~Li \emph{et~al.}, ``Vila-mil: Dual-scale vision-language multiple instance learning for whole slide image classification,'' in \emph{CVPR}, 2024, pp. 11\,248--11\,258.

\bibitem{seyfioglu2024quilt}
M.~S. Seyfioglu, W.~O. Ikezogwo \emph{et~al.}, ``Quilt-llava: Visual instruction tuning by extracting localized narratives from open-source histopathology videos,'' in \emph{CVPR}, 2024, pp. 13\,183--13\,192.

\bibitem{sun2024pathgen}
Y.~Sun, Y.~Zhang \emph{et~al.}, ``Pathgen-1.6 m: 1.6 million pathology image-text pairs generation through multi-agent collaboration,'' \emph{arXiv preprint arXiv:2407.00203}, 2024.

\bibitem{sun2024cpath}
Y.~Sun, Y.~Si \emph{et~al.}, ``Cpath-omni: A unified multimodal foundation model for patch and whole slide image analysis in computational pathology,'' \emph{arXiv preprint arXiv:2412.12077}, 2024.

\bibitem{resnet}
K.~He, X.~Zhang \emph{et~al.}, ``Deep residual learning for image recognition,'' in \emph{CVPR}, 2016, pp. 770--778.

\bibitem{vit}
A.~Dosovitskiy, L.~Beyer \emph{et~al.}, ``An image is worth 16x16 words: Transformers for image recognition at scale,'' in \emph{ICLR}, 2021.

\bibitem{gcn}
T.~N. Kipf and M.~Welling, ``Semi-supervised classification with graph convolutional networks,'' \emph{arXiv preprint arXiv:1609.02907}, 2016.

\bibitem{RRTMIL}
W.~Tang, F.~Zhou \emph{et~al.}, ``Feature re-embedding: Towards foundation model-level performance in computational pathology,'' in \emph{CVPR}, June 2024, pp. 11\,343--11\,352.

\bibitem{UBC-OCEAN}
M.~Asadi-Aghbolaghi, H.~Farahani \emph{et~al.}, ``Ubc ovarian cancer subtype classification and outlier detection (ubc-ocean),'' \url{https://kaggle.com/competitions/UBC-OCEAN}, 2023, kaggle.

\bibitem{howard2023integration}
F.~M. Howard, J.~Dolezal \emph{et~al.}, ``Integration of clinical features and deep learning on pathology for the prediction of breast cancer recurrence assays and risk of recurrence,'' \emph{NPJ Breast Cancer}, vol.~9, no.~1, p.~25, 2023.

\bibitem{salonia2023eau}
A.~Salonia, C.~Bettocchi \emph{et~al.}, ``Eau guidelines,'' in \emph{The EAU Annual Congress Milan; EAU Guidelines Office: Arnhem, The Netherlands}, 2023.

\bibitem{gpt4}
J.~Achiam, S.~Adler \emph{et~al.}, ``Gpt-4 technical report,'' \emph{arXiv preprint arXiv:2303.08774}, 2023.

\bibitem{dao2023flashattention2}
T.~Dao, ``Flash{A}ttention-2: Faster attention with better parallelism and work partitioning,'' in \emph{ICLR}, 2024.

\bibitem{gat}
P.~Veli{\v{c}}kovi{\'c}, G.~Cucurull \emph{et~al.}, ``Graph attention networks,'' \emph{arXiv preprint arXiv:1710.10903}, 2017.

\bibitem{graphsage}
W.~Hamilton, Z.~Ying \emph{et~al.}, ``Inductive representation learning on large graphs,'' \emph{NeurIPS}, vol.~30, 2017.

\bibitem{team2023gemini}
G.~Team, R.~Anil \emph{et~al.}, ``Gemini: a family of highly capable multimodal models,'' \emph{arXiv preprint arXiv:2312.11805}, 2023.

\bibitem{claude3}
Anthropic. (2024) Claude 3 haiku: our fastest model yet.

\bibitem{llama}
H.~Touvron, T.~Lavril \emph{et~al.}, ``Llama: Open and efficient foundation language models,'' \emph{arXiv preprint arXiv:2302.13971}, 2023.

\bibitem{gpt3}
T.~B. Brown, ``Language models are few-shot learners,'' \emph{arXiv preprint ArXiv:2005.14165}, 2020.

\end{thebibliography}
\end{document}